\PassOptionsToPackage{dvipsnames}{xcolor}

\documentclass{article}

\usepackage{microtype}
\usepackage{hyperref}
\usepackage{url}
\usepackage{booktabs}
\usepackage{graphicx}
\usepackage{xcolor}
\usepackage{subcaption}
\usepackage[most]{tcolorbox}
\usepackage{enumitem}
\usepackage{CJKutf8}
\definecolor{softblue}{rgb}{0.88, 0.95, 1.0} 
\definecolor{softyellow}{rgb}{0.98, 0.98, 0.82} 

\usepackage{multirow}
\usepackage{lineno}
\usepackage{wrapfig}

\usepackage[utf8]{inputenc} 
\usepackage[T1]{fontenc}    
\usepackage{hyperref}       
\usepackage{url}            
\usepackage{booktabs}       
\usepackage{amsfonts}       
\usepackage{nicefrac}       
\usepackage{microtype}      
\usepackage{graphicx}
\usepackage{float}
\usepackage{natbib}
\usepackage{doi}
\usepackage{xcolor}
\usepackage{lineno}
\usepackage{amsmath}
\usepackage{xcolor}
\usepackage{enumitem}
\usepackage{arxiv}
\usepackage{url}
\newcommand\myshade{85}
\colorlet{mylinkcolor}{RoyalBlue}
\colorlet{mycitecolor}{violet}
\colorlet{myurlcolor}{YellowOrange}

\hypersetup{
  linkcolor  = mylinkcolor!\myshade!black,
  citecolor  = mycitecolor!\myshade!black,
  urlcolor   = myurlcolor!\myshade!black,
  colorlinks = true,
}

\usepackage{color}

\usepackage{times}
\usepackage{latexsym}

\usepackage[T1]{fontenc}

\usepackage[utf8]{inputenc}

\usepackage{microtype}

\usepackage{booktabs}
\usepackage{graphicx}
\usepackage{makecell}
\usepackage{multirow}
\usepackage{longtable}
\usepackage{CJKutf8}
\usepackage{float}
\usepackage{algpseudocode}
\usepackage[frozencache, cachedir=minted-cache]{minted}

\usepackage{tikz}
\usetikzlibrary{shapes.geometric}

\usepackage{wrapfig}
\usepackage{cleveref}
\usepackage[ruled]{algorithm2e}
\crefname{algocf}{alg.}{algs.}

\title{
Dynamic Generation of Personalities with Large Language Models
}

\newcommand*\samethanks[1][\value{footnote}]{\footnotemark[#1]}

\author{
\small
    Jianzhi Liu\textsuperscript{1}\quad 
    Hexiang Gu\textsuperscript{1}\quad 
    Tianyu Zheng\textsuperscript{1}\quad\\
\textbf{    Liuyu Xiang\textsuperscript{1}\quad
    Huijia Wu\textsuperscript{1}\quad
    Jie Fu\textsuperscript{2}\begin{NoHyper}\thanks{Corresponding author.}\end{NoHyper}\quad 
    Zhaofeng He\textsuperscript{1}\samethanks[1]\quad}
\\
\small
    \textsuperscript{1}Beijing University of Posts and Telecommunications\quad \textsuperscript{2}HKUST\quad 
\\
\small
\texttt{\{lujianzhi,guhexiang,zhengtianyu,liuyuxiang,huijiawu,zhaofenghe\}@bupt.edu.cn} \quad \\
\texttt{jiefu@ust.hk}
}

\begin{document}

\maketitle

\begin{abstract}
In the realm of mimicking human deliberation, large language models (LLMs) show promising performance, thereby amplifying the importance of this research area. Deliberation is influenced by both logic and personality. However, previous studies predominantly focused on the logic of LLMs, neglecting the exploration of personality aspects. In this work, we introduce Dynamic Personality Generation (DPG), a dynamic personality generation method based on Hypernetworks. Initially, we embed the Big Five personality theory into GPT-4 to form a personality assessment machine, enabling it to evaluate characters' personality traits from dialogues automatically. We propose a new metric to assess personality generation capability based on this evaluation method. Then, we use this personality assessment machine to evaluate dialogues in script data, resulting in a personality-dialogue dataset. Finally, we fine-tune DPG on the personality-dialogue dataset. Experiments prove that DPG's personality generation capability is stronger after fine-tuning on this dataset than traditional fine-tuning methods, surpassing prompt-based GPT-4.
\end{abstract}

\section{Introduction}

Large language models (LLMs) such as GPT-4~\citep{gpt-4} and Palm~\citep{palm} demonstrate impressive cognitive abilities. This enables LLMs to deliberate like humans. However, human deliberation is influenced by both logic and personality~\citep{cantor1990thought,griffin2015cognition}. Previous research primarily focuses on enabling LLMs to simulate human deliberation from a logical perspective~\citep{ghost,lytMC,chen2023agentverse}, while neglecting the study of personality aspects. Exploring how to shape the personalities of LLMs and influence their deliberation is a topic worth studying.

In psychological research, the Big Five personality traits are recognized as models for studying human personality~\citep{costa1992five}. Some works introduce them into LLM research as a method to measure the personalities of LLMs~\citep{pt7,safdari2023personality}. To impart specific personality traits to pre-trained language models requires complex and well-designed prompts~\citep{li2023qinggan,jiang2023qinggan}. The proficiency of prompt designers significantly influences the accuracy of personality shaping~\citep{safdari2023personality}. Some approaches address the issue of heavy prompt dependency by incorporating external knowledge bases or utilizing supervised fine-tuning with personalized data~\citep{ wang2023rolellm,li2023chatharuhi}. However, these methods introduce data dependencies, as shaping a new personality necessitates gathering a specific set of role-specific data.
Therefore, the challenge lies in creatively generating new personalities while reducing reliance on prompts and data.

\begin{figure}[htbp]

\begin{center}
\includegraphics[scale=0.3]{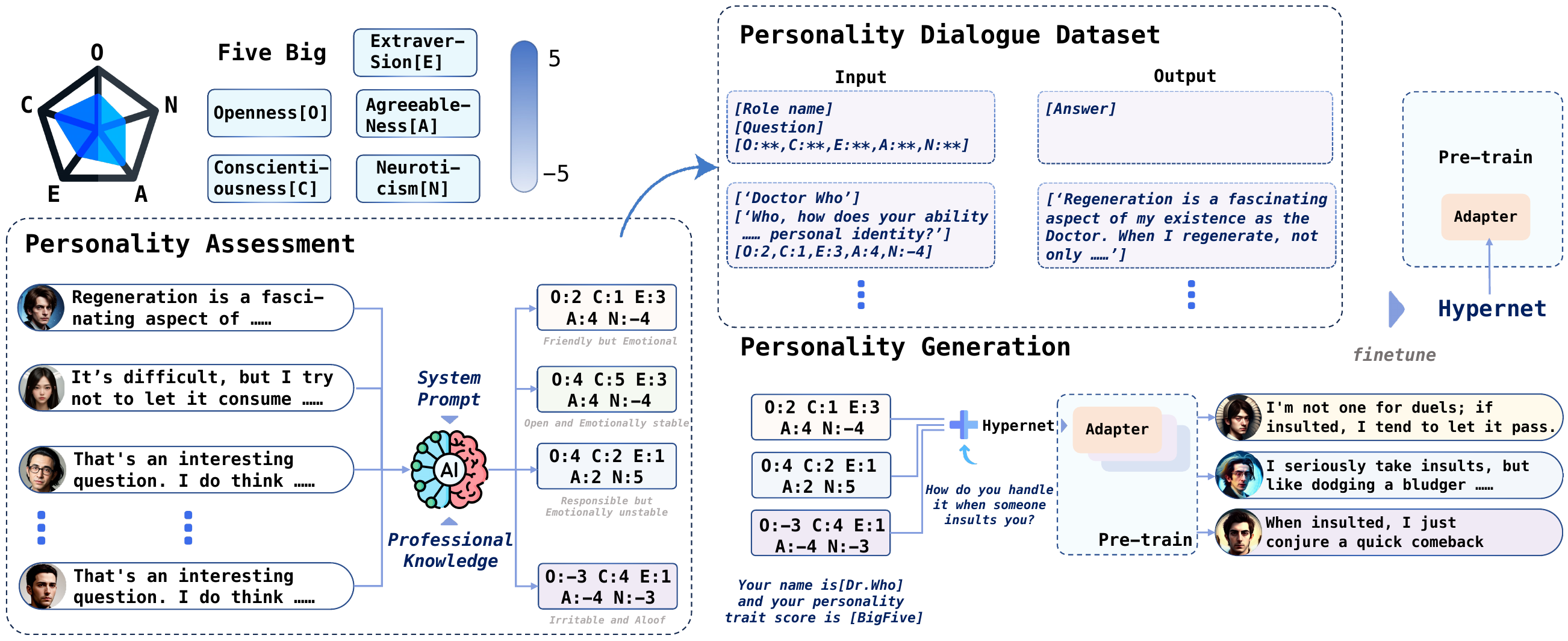}
\end{center}
\caption{Illustration of the Dynamic Generation of Personalities (DPG). \textbf{Personality Assessment:}The Big Five personality traits are quantified into 11 scores ranging from -5 to 5. GPT-4, equipped with expertise in personality assessment, evaluates the character's Big Five personality traits through dialogue. \textbf{Personality Generation:} Adapters are inserted into the pre-trained LLMs, and hypernetworks are trained using dialogue data with personality scores. This allows for the generation of different adapter weights based on the prompt, enabling the LLMs to exhibit diverse personalities.}
\label{main_figure}
\end{figure}

In this paper, we propose a novel approach to personality generation called Dynamic Personality Generation (DPG), as depicted in Figure~\ref{main_figure}. Our goal is to reshape the personalities of LLMs by integrating the personality traits and dialogue styles of different script characters. Drawing upon psychological knowledge, we employ GPT-4 to generate personality trait markers and annotate a dataset of personality-character dialogues. Subsequently, we train Hypernetworks~\citep{ha2016hypernetworks} to generate adapters for LLMs, effectively shaping their personalities. Experimental results demonstrate the effectiveness of our annotated data in personality shaping for LLMs. Our DPG approach outperforms traditional fine-tuning methods and state-of-the-art pre-trained models like GPT-4, offering a more dynamic and nuanced approach to personality shaping in large-scale models.

This work makes the following contributions:
\begin{itemize}[leftmargin=20pt]
\item [$\bullet$]
To assess personality traits from role dialogues, we improve existing machine assessment methods~\citep{pt7} by incorporating knowledge from the Big Five personality traits. This approach allows us to analyze character dialogues and generate more stable and reliable assessments of personality traits. 
\item [$\bullet$]
We evaluate the personality traits of characters involved in conversations through dialogues from novels, movies, and other scripts, creating a personality-dialogue dataset. Experiments prove the effectiveness of this dataset in generating characters with personalities.
\item [$\bullet$]
We propose a method called Dynamic Personality Generation (DPG) that enables the generation of new personalities in LLMs by integrating personality information from historical dialogues. Experimental results demonstrate that DPG outperforms traditional fine-tuning methods and prompt-based GPT-4~\citep{gpt-4}. 
\end{itemize}


\section{Task Formulation}
The research aims to develop a methodology for the LLM chatbot to adopt different personalities based on user prompts. This involves evaluating conversations to ensure the generated personality aligns with the prompt and engages in meaningful and coherent conversations.

\section{Related Work}\label{Related work}

\textbf{Personality Assessment for Large Language Models:}
The lexical hypothesis suggests that it's possible to assess an individual's personality through dialogue~\citep{costa1992five,fivebig2,chjs,fivebig3}. Recent work evaluates the Big Five personality traits and Myers-Briggs Type Indicator of large language models using open-ended questions and interviews questionnaires~\citep{pt1,pt2,pt3,pt4,pt5,pt6}. Using tools such as Apply Magic Sauce\footnote{https://applymagicsauce.com/demo} and GPT-4 to predict personality traits from conversations proves to be effective\citep{pt7,safdari2023personality}.

\textbf{Personality Editing of Large Language Models:}
Recent studies showcase the remarkable capabilities of LLMs for generating distinct personalities. These models achieve this by integrating external knowledge bases~\citep{rpknow,rpknow2}, employing complex multi-turn prompts~\citep{rpprompt}, and leveraging unique script data through fine-tuning. By adopting role-playing or personality-driven approaches, these models can assume specific personalities~\citep{rp1,rp2,rp3,rp4,rp5,rp6,rp7,rp8,rp9,safdari2023personality}. However, these methods have limitations in terms of reconstructing existing characters and cannot generate new ones with creativity. Additionally, each character necessitates a separate and detailed dataset for external knowledge or fine-tuning. Therefore, our research is dedicated to developing a method that effectively utilizes script data to generate new personalities.

\section{Data Construction}\label{dataengine}

Our research greatly benefits from the contributions of ROLELLM~\citep{wang2023rolellm} and ChatHaruhi~\citep{li2023chatharuhi}. 
They provide us with character dialogue data that spans a wide range of personality traits, represented as ${D_{Script}, D_{Augment}}$.
In this case, $D_{Script}$ represents real character dialogues extracted from novels and other texts, while $D_{Augment}$ represents character-generated dialogues enhanced by LLMs like GPT-4.

\begin{figure}[h]
  \begin{minipage}{0.49\textwidth}
    \centering
    \small
    \renewcommand{\arraystretch}{1.25}
    \begin{tabular}{lc} 
    \toprule
      Key & Value \\ 
      \midrule
      Dialogue rounds(Part/All) & \\
      \quad - In English & 16896/56547 \\
      \quad - In Chinese & 39649/56547 \\ 
      \midrule
      Big Five Personality(High/Low) & \\
      \quad Openness & \\
      \quad \quad -In English & 15458/1438 \\
      \quad \quad -In Chinese & 36829/2820 \\
      \quad Conscientiousness & \\
      \quad \quad -In English & 14112/2784 \\
      \quad \quad -In Chinese & 33734/5915 \\
      \quad Extroversion & \\
      \quad \quad -In English & 12264/4632 \\
      \quad \quad -In Chinese & 33599/6050 \\
      \quad Agreeableness & \\
      \quad \quad -In English & 9359/7537 \\
      \quad \quad -In Chinese & 22106/17543 \\
      \quad Neuroticism & \\
      \quad \quad -In English & 6683/10213 \\
      \quad \quad -In Chinese & 7794/31855 \\
    \bottomrule
    \end{tabular}
    \captionof{table}{Statistics of the Dataset}
    \label{table:dataset}
  \end{minipage}%
\hfill
  \begin{minipage}{0.49\textwidth}
  \vspace{0.15cm}
    \centering
    \vspace{0.15cm}
    \includegraphics[width=0.8\textwidth]{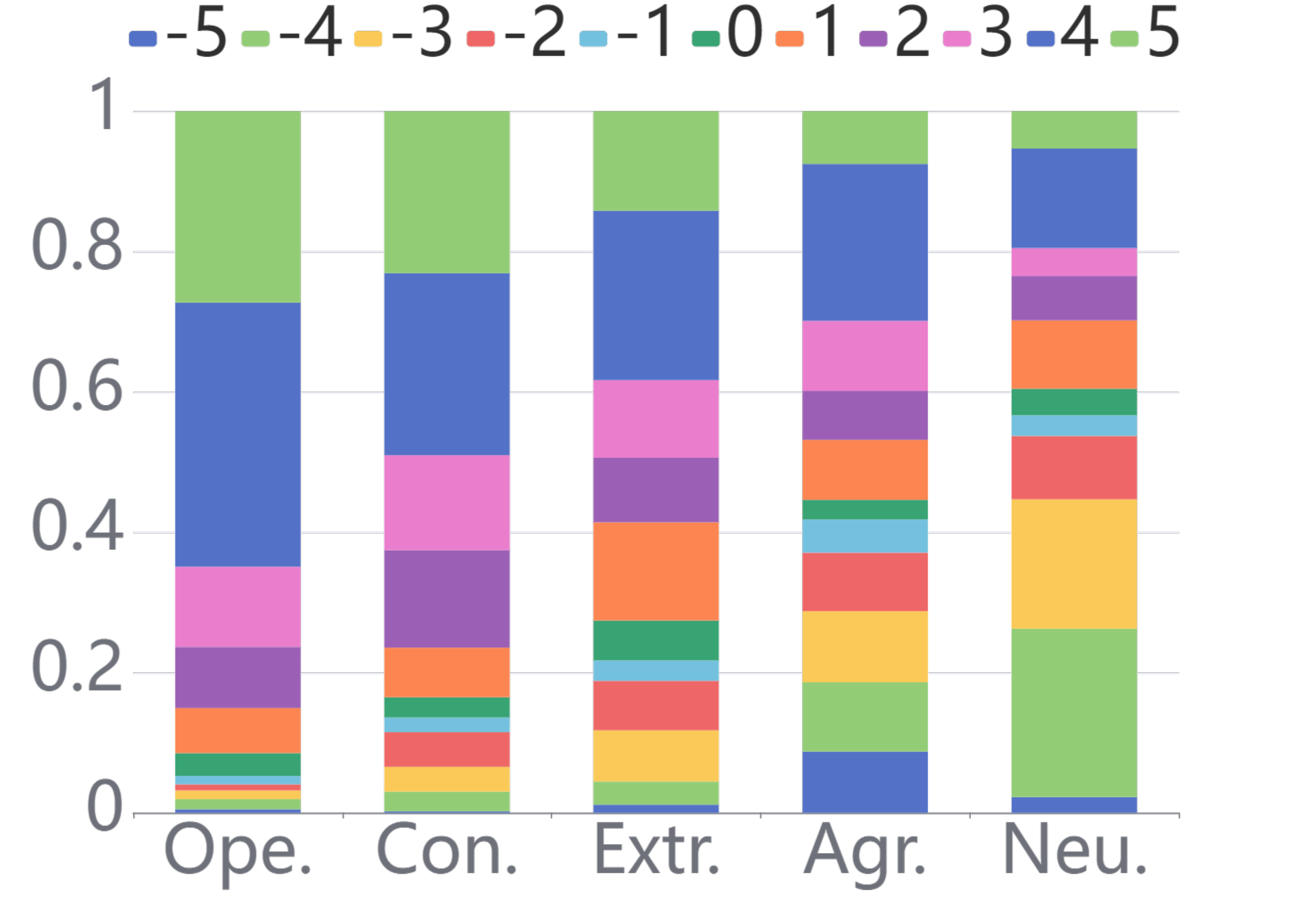}\\
    \vspace{0.15cm}
    \textbf{(a)} English Dataset
    \vspace{0.1cm} 
    \includegraphics[width=0.8\textwidth]{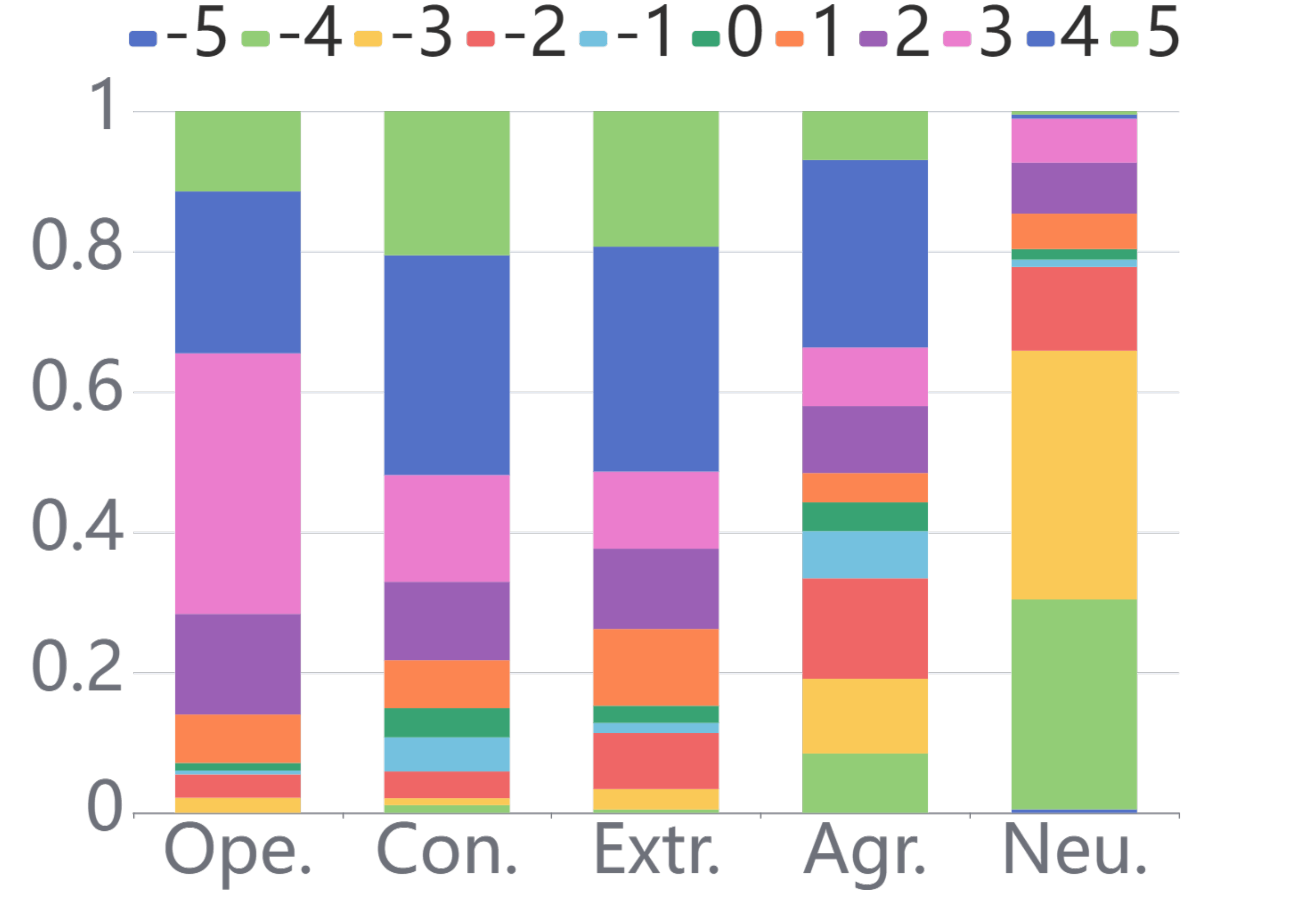}\\
    \vspace{0.1cm}
    \textbf{(b)} Chinese Dataset
    \vspace{0.15cm}
    \caption{Personality Rating Statistics, different colors represent distinct scores.}
    \label{fig:datasets}
  \end{minipage}
\end{figure}

In this section, we introduce personality assessment and data cleaning work. Firstly, we clarify the lack of coherence and completeness in the dialogue data (Section~\ref{3.1}).
We then test the stability of three personality evaluators to obtain a more reliable personality assessment method (Section~\ref{3.2}). 
Lastly, we examine the personality-dialogue dataset and remove data points where the personality traits significantly deviate from the character's central personality (Section~\ref{3.3}). 
As a result, we obtain a clean personality-dialogue dataset that is suitable for personality generation. The statistical information of the dataset is presented in Figure~\ref{fig:datasets} and Table~\ref{table:dataset}.

\subsection{The Coherence and Contextual Integrity of the Dialogues}
\label{3.1}
The performance of LLMs in role-playing tasks can be negatively affected if they are fine-tuned with data that contains noise~\citep{wang2023rolellm}. 
This is because LLMs tend to internalize and reproduce the inconsistencies and irregularities present in the noisy training data. 
As a result, the models may struggle to generate coherent and contextually appropriate responses, diminishing their effectiveness in role-playing scenarios.
It is crucial to prioritize the cleanliness and relevance of training data to ensure the optimal performance and proficiency of LLMs in complex tasks like role-playing. The effectiveness of GPT as an evaluator has been firmly established in previous studies~\citep{gptky1,gptky2,gptky3}. We utilize the evaluation capabilities of GPT-3.5 to assess the \textbf{Coherence} and \textbf{Contextual Integrity} of dialogues. 

\textbf{Coherence:} Responses should be relevant to the questions, and each dialogue turn should revolve around the same topic.

\textbf{Contextual Integrity: }Dialogues should include sufficient contextual information to ensure that responses are generated based on the questions. 

\subsection{The Reliability of Personality Assessment}
\label{3.2}
\subsubsection{The Big Five Personality}
The Big Five Personality Traits is a widely accepted model in psychology that describes the five core dimensions of human personality, which are \textbf{Openness(O)}, \textbf{Conscientiousness(C)}, \textbf{Extraversion(E)}, \textbf{Agreeableness(A)}, and \textbf{Neuroticism(N)} \citep{costa1992five}. 

\subsubsection{Confidence in Personality Assessment Methods}

Evaluating personality traits from dialogue data using machines has been proven effective~\citep{pt7,safdari2023personality}. We map each dimension of the Big Five personality traits to integers ranging from -5 to 5, similar to the approach in~\citep{pt7}. Lower scores indicate lower attributes of that personality, while higher scores indicate higher attributes. We then enhance the LLMs-based personality assessment machine in~\citep{pt7} by incorporating expert knowledge of the Big Five personality traits~\citep{costa1992five}, resulting in \textbf{GPT-4 with Embedded Psychological Knowledge $(F_{Knowledge-gpt4})$}. 
Internal consistency is a vital indicator used to evaluate the reliability of measuring the Big Five personality traits~\citep{xdxs1,xdxs2,xdxs3,xdxs4}. Similar to Jian~\citep{jiang2024evaluating}, we employ variance as the foundation for assessing the effectiveness of the following three methods: \textbf{Prompt-based GPT-3.5} $(F_{Prompt-gpt3.5})$, \textbf{Prompt-based GPT-4 $(F_{Prompt-gpt4})$} from~\citep{pt7}, and \textbf{GPT-4 with Embedded Psychological Knowledge $(F_{Knowledge-gpt4})$} which we have enhanced. For more detailed information on the assessment process, please refer to Appendix~\ref{knowledge}.
\begin{table}[htbp]

\begin{center}
\begin{tabular}{lllllll}
\toprule
\multicolumn{1}{c}{\bf Method}  &\multicolumn{1}{c}{\bf $\sigma_{O}^{2} $} & \bf$\sigma_{C}^{2} $& \bf $\sigma_{E}^{2} $& \bf $\sigma_{A}^{2} $&\bf $\sigma_{N}^{2} $ &$\sigma_{Avg.}^{2} $\\
\midrule
$F_{Prompt-gpt3.5}$&0.940& 0.794&  1.199& 1.344& 0.869&1.0292
\\
$F_{Prompt-gpt4}$&0.728&  0.673&  0.689& \colorbox{softblue}{\textbf{1.210}}& 0.887&0.8374
\\
$F_{Knowledge-gpt4}$& \colorbox{softblue}{\textbf{0.487}}& \colorbox{softblue}{\textbf{0.659}}& \colorbox{softblue}{\textbf{0.633}}& 1.249& \colorbox{softblue}{\textbf{0.692}}&\colorbox{softblue}{\textbf{0.744}}\\
\bottomrule
\end{tabular}
\end{center}
\caption{The intrinsic consistency evaluation results of \textbf{Prompt-based GPT-3.5}, \textbf{Prompt-base 
 GPT-4} and \textbf{GPT-4 with Embedded Psychological Knowledge}. $\sigma_{(O|C|E|A|N)}^{2}$ represents the variance in assessments of openness, conscientiousness, extraversion, agreeableness, and neuroticism.  $\sigma_{Avg.}^{2}$ denotes the average variance. The results indicate that $F_{Knowledge-gpt4}$ demonstrates superior internal consistency in assessments other than agreeableness compared to other methodologies.  The results of our best model are in
\colorbox{softblue}{Blue}.}
 \label{fangcha}
\end{table}

Table \ref{fangcha} presents the evaluation results, indicating that $F_{Knowledge-gpt4}$ provides a more reliable assessment of personality traits from dialogues. 
Following the same experimental procedure outlined earlier, we employed $F_{Knowledge-gpt4}$ to evaluate our entire dialogue dataset, resulting in the creation of a dataset denoted as $D={D_{ScriptPair}, D_{AugmentPair}}$. $D_{ScriptPair}$ and $D_{AugmentPair}$ are represented as ${<dialogueSet, (O, C, E, A, N)>}$, where $dialogueSet$ represents a set of dialogues, and $(O, C, E, A, N)$ represents the Big Five personality assessment scores for the evaluated character in that set.

\subsection{Personality Displacement Caused by LLMs Enhancement}
\label{3.3}
We present the personality assessment scores in a coordinate space, revealing that each character's personality revolves around a central point, which we refer to as the personality center. 
This observation is consistent with the conclusion drawn in~\citep{pt7}. 
The partiality of personality assessment based on dialogue likely contributes to this pattern.
 
There are concerns that enhanced data compared to script data may produce excessive personality biases. 
For characters with both script data and enhanced data, we calculated the mean \(\mu_s, \mu_g\) and variances \(\sigma^2_s, \sigma^2_g\) for the assessment scores \((O, C, E, A, N)_{Script}\) and \((O, C, E, A, N)_{Augment}\). 
We found that the personality center of the assessment displacement after enhancement (\(\mu_s\) differs \(\mu_g\)).
Additionally, characters in the script data exhibit a more diverse personality compared to those in the enhanced data (\(\mu_{\sigma^2_s}>\mu_{\sigma^2_g}\)). 
To reduce the impact of outlier data on training, we removed data that deviated excessively from the personality center. (See Appendix~\ref{knowledge} for details.)

\section{Network Method}\label{others}

\subsection{Personality Shaping Based on LoRA}
\textbf{LoRA}~\citep{hu2021lora} is developed to train extensive pre-trained models while minimizing the need for training resources.
Previous research demonstrates the effectiveness of utilizing LoRA for fine-tuning LLMs for downstream tasks.
In our study, we employ LoRA-based fine-tuning on the personality-dialogue dataset personality dataset to train our LLMs.
This approach allows LoRA to imbue LLMs with specific personality traits, thereby shaping the model's personality.

To achieve this, we apply LoRA to each Attention layer of LLMs, utilizing the following mathematical formula~\ref{eq4}:

\begin{equation}
    \label{eq4}y = F_{\theta _{Pre-train}}(x)+F_{\theta _{LoRA_B}}(F_{\theta _{LoRA_A}}(x))
\end{equation}

Here, $x$ represents the input of the attention layer, $y$ denotes the output of the attention layer, and $\theta$ signifies the neural network parameters. $\theta_{Pre-train}$ represents the weights of the pre-trained layer.
In this context, the parameters $\theta_{LoRA_A}$ and $\theta_{LoRA_B}$ play a crucial role in shaping the personality of LLMs.

\subsection{Personality Generation Based on Hypernetworks}

\begin{figure}[h]
\begin{center}
\includegraphics[scale=0.25]{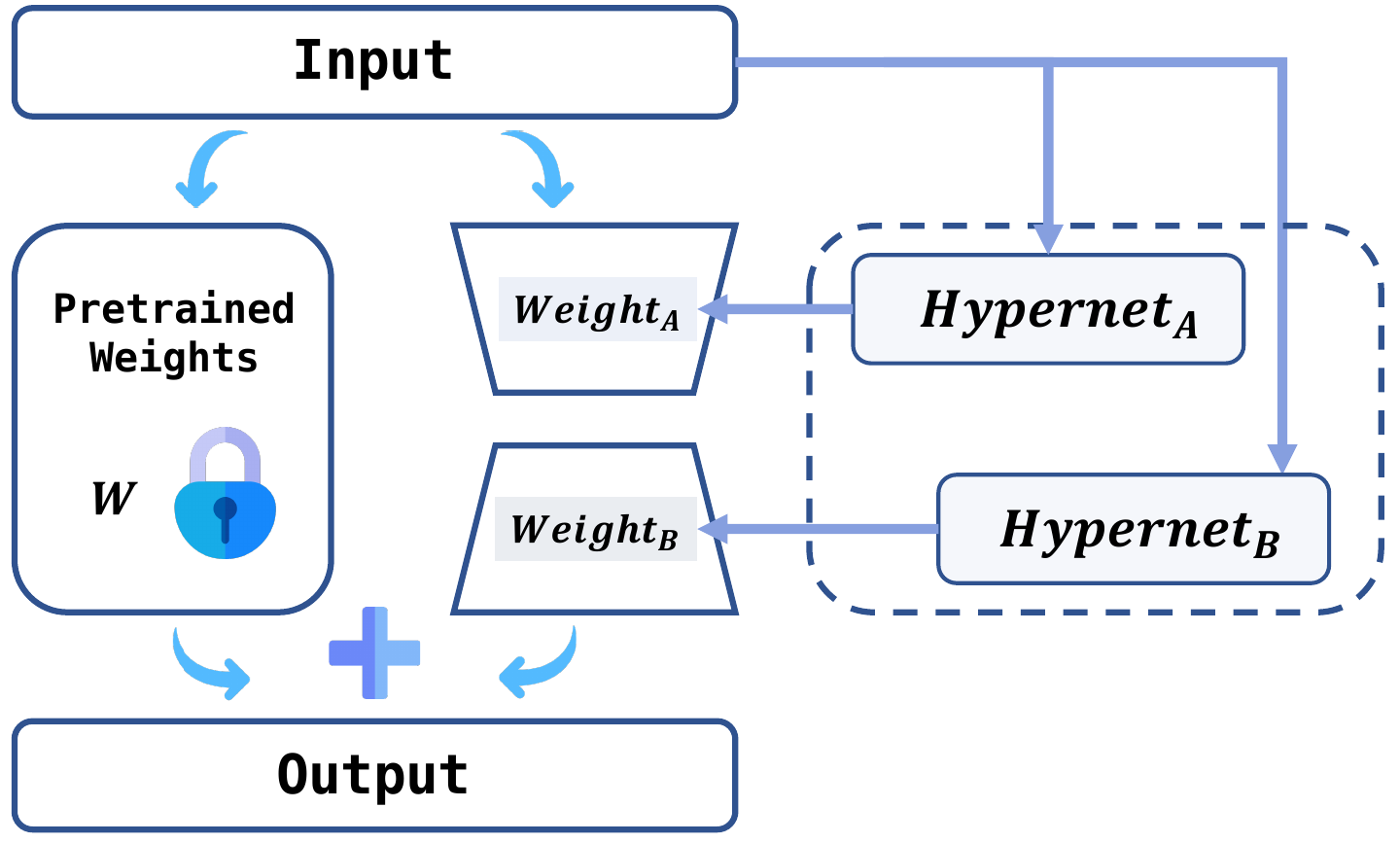}
\end{center}
\caption{Generating Lora weights based on Hypernetworks. 
The Pretrained weights $W$ are frozen (\protect\includegraphics[width=0.3cm]{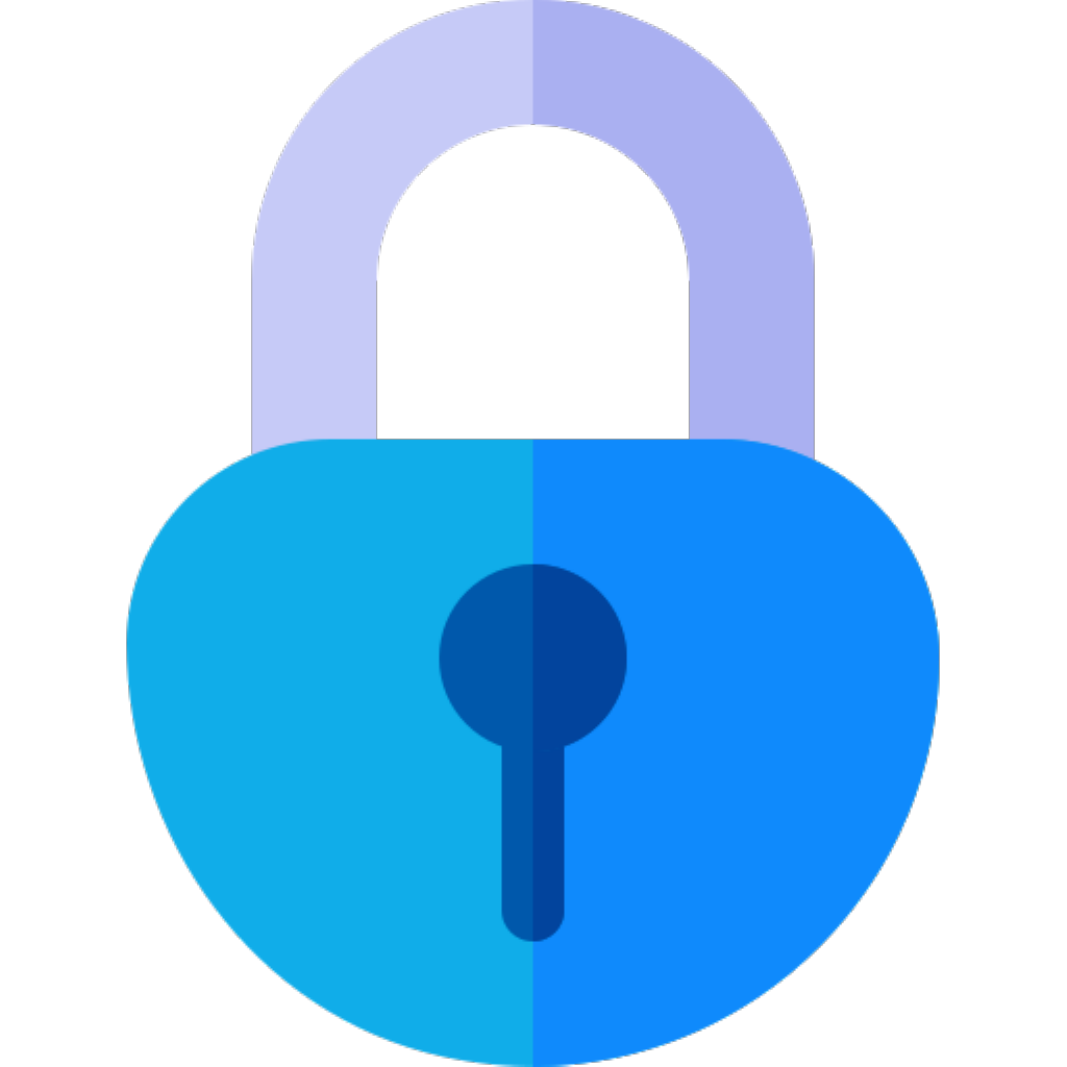}), and the LoRA's weights ($Weight_{A}$ and $Weight_{B}$) are generated by hypernet ($Hypernet_{A}$ and $Hypernet_{B}$), with only the weights of the hypernet being trained. \protect\includegraphics[width=0.3cm]{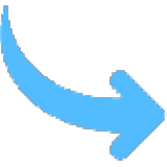} represents the data transfer path of LoRA~\citep{hu2021lora}. \protect\includegraphics[width=0.3cm]{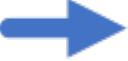} represents the data transfer path where hypernet generates LoRA's weights}
\label{hyperlora}
\end{figure}

\textbf{Hypernetworks}~\citep{ha2016hypernetworks} is a technique that allows one neural network to generate parameters for another neural network within the same architecture.
This approach enhances the flexibility of the network, enabling it to dynamically adjust its network weights based on various tasks or input data.
In our study, we employ Hypernetworks as the generator for the weights of the LoRA layer, as depicted in Figure~\ref{hyperlora}. 
By using Hypernetworks, we can generate different sets of LoRA weights for different inputs, representing distinct personality traits. This generative capability empowers the LLMs to produce content with various personality traits. The mathematical formulas are presented in Equation \ref{eq5}.
\begin{equation}
    \label{eq5}\theta _{LoRA_{*}} = F_{Hypernet_{*}}(x) 
\end{equation}
Where $\theta_{LoRA_{*}}$ represents the weights of the LoRA layer, carrying personality traits. $F_{Hypernet_{*}}$ denotes  MLP-based Hypernetworks neural networks used for generative personality trait expression. $x$ represents the input to the network.

\subsection{Implementation of LoRA Based on Hypernetworks in LLMs}
We integrate a Hypernetworks-based LoRA layer structure into each Attention layer of the LLMs. This guarantees that our approach is plug-and-play compatible with pre-trained LLMs of various transformer architecture sizes. The parameters of the pre-trained model are kept frozen, and we exclusively train the Hypernet network within each layer. Under this network architecture, we can conduct end-to-end training without modifying the parameters of the pre-trained model, thus allowing the LLMs to develop distinct personality traits.

\section{Experiments}

\subsection{Experiment Setup}
\textbf{Dataset: }We conduct experiments using a personality-dialogue dataset. The Chinese and English data are split and used for parallel experiments without interfering with each other. The dataset is preprocessed into a supervised training format, as described in Appendix \ref{traindata}.

\textbf{Base Models:} For the Chinese data, we use Glm3-3B-Chat~\citep{glm3} and Yi-6B-Chat~\citep{yi} as the pre-training bases. As for the English data, we use LLama2-7B-Chat-hf~\citep{llama2} and Yi-6B-Chat~\citep{yi} as the pre-training bases.

\textbf{Baseline:} Our baseline consists of three distinct approaches: 
\vspace{-0.1cm}
\begin{itemize}[leftmargin=20pt]
\item [$\bullet$]
Prompt-based pre-trained model. Here, GPT-4~\citep{gpt-4} is utilized as the pre-trained model. Two versions of prompts are used, including one with a system (GPT-4 w/i Sys.P.) prompt and one without (GPT-4 w/o Sys.P.).
\item [$\bullet$]
Applying supervised fine-tuning to a subset of the parameters of the pre-trained model (Freeze-SFT).~\citep{llama-factory} provides support for this method. 
\item [$\bullet$]
Implementing LoRA-based\citep{hu2021lora} supervised fine-tuning on the pre-trained model (LoRA-SFT).~\citep{peft} provides support for this method. 
\end{itemize}

\textbf{Evaluation Criteria:} Conversational ability and personality-shaping ability are used as evaluation criteria.
\vspace{-0.1cm}
\begin{itemize}[leftmargin=20pt]
\item [$\bullet$]
Conversational Ability: We are inspired by CharacterEval\citep{bm1} and use three metrics for evaluation: \textbf{Fluency (Flu.)}, \textbf{Coherency (Coh.)}, and \textbf{Consistency (Con.)} to ensure readable and criteria-compliant answers. Prompt-based GPT-3.5 is used as an evaluator (refer to Appendix\ref{eval_prompt}).
\item [$\bullet$]
Personality Shaping Ability: We employ the Euclidean distance between the prompted personality and the evaluated personality as the \textbf{Personality Shaping Deviation Coefficient (P.S.D.C.)} to assess the ability to shape personality. 
\end{itemize}
\textbf{Test Methods:} 
\vspace{-0.1cm}
\begin{itemize}[leftmargin=20pt]
\item [$\bullet$] For pre-trained models, we use the [character name],  [character prompt personality], and [the five historical character dialogues closest to the prompt personality for that character from the dataset] as the system prompt. We use the test question as the user prompt for testing.
\item [$\bullet$] For Freeze-SFT, LoRA-SFT, and DPG, we first fine-tune the model using the dataset and then use the character name, character prompt personality, and test question as the prompts for testing.
\end{itemize}

Section~\ref{valset},~\ref{openq}, and~\ref{pg} introduce three testing tasks, respectively.

\subsubsection{Performance in Scripted Dialogue}
\label{valset}
We select ten characters with the highest number of dialogue rounds from the personality dialogue dataset to serve as [character names]. For each character, we randomly selected five dialogue excerpts (including personality ratings [prompt personality] and dialogue content [questions for testing]) as the test set. The remaining data serves as the training set to train our model and baseline. Additionally, for pre-training, we also need to select five additional dialogue excerpts as supplementary information for the character profiles.

\subsubsection{Performance Evaluation using Open-Ended Questions }
\label{openq}
To further evaluate the performance of personality shaping, we use the open-ended questions provided in~\citep{pt7} as the test set. For each character, their personality central tendency (the average of personality ratings) serves as the prompt personality. The remaining evaluation settings are the same as in~\ref{valset}.

\subsubsection{Evaluate the Ability to Generate New Personality  }
\label{pg}
To evaluate the model's ability to generate entirely new personalities, we rate the Big Five personality traits of openness, neuroticism, and agreeableness on four levels: -5, -2, 2, and 5. Regarding conscientiousness and extraversion, given the significant deviation of the model from average human values~\citep{pt7,safdari2023personality}, we opt to calculate their mean as a substitute (Conscientiousness mean: 3, Extraversion mean: 2). By arranging and combining these five dimensions of traits, we form 64 unique personality combinations. These personality combinations serve as replacements for the prompt in~\ref{openq} personalities during the experiments.

\subsection{Main Result}
The experimental results are presented using LLama-7B as an example. For more experimental results and detailed descriptions, refer to Appendix \ref{detailed assessment}.

\begin{table}[htbp]
\centering
\begin{tabular}{ccccc}
\toprule
\multicolumn{1}{c}{\textbf{Method}} & Flu.$\uparrow$ & \bf Coh.$\uparrow$ & \bf Con.$\uparrow$ & \bf P.S.D.C.$\downarrow$ \\ \midrule
Freeze-SFT                          &  84.31\% & 72.23\% & 53.14\% & 5.52  \\
LoRA-SFT                            &  97.96\% & 94.31\% & 91.12\% & 4.25   \\
DPG(ours)                      & 98.39\% &\colorbox{softblue}{\textbf{95.64\%}} & \colorbox{softblue}{\textbf{92.23\%}} & \colorbox{softblue}{\textbf{3.14}} \\ \hline \midrule
GPT-4  w/o Sys.P.                &  \colorbox{softblue}{\textbf{99.73\%}} &86.13\% &84.67\% &3.44  \\
GPT-4  w/i Sys.P.                    &99.31\% &87.48\% &85.21\% &3.36  \\
\bottomrule
\end{tabular}
\caption{
The evaluation results of the \textbf{Script Dialogue Test} for Llama-7B-En. Fluency (Flu.), Coherence (Coh.), and Consistency (Con.) represent three dimensions of conversational ability, while the personality Shaping Deviation Coeﬀicient
(P.S.D.C.) represents the capability to shape personality. The data (\%) indicates the proportion of cases considered correct out of the total test cases. The results of our best model are in
\colorbox{softblue}{Blue}.}
\label{script-test}
\end{table}

\begin{table}[htbp]
\centering
\begin{tabular}{ccccc}
\toprule
\multicolumn{1}{c}{\textbf{Method}} & Flu.$\uparrow$ & \bf Coh.$\uparrow$ & \bf Con.$\uparrow$ & \bf P.S.D.C.$\downarrow$ \\ \midrule
Freeze-SFT                          & 73.47\%& 42.16 \%&21.84\% &6.23  \\
LoRA-SFT                            & 88.27\%&87.97\% & 84.67\%&4.92   \\
DPG(ours)       & 94.09\%&\colorbox{softblue}{\textbf{93.27\%}}&\colorbox{softblue}{\textbf{91.06\%}}&\colorbox{softblue}{\textbf{2.86}} \\ \hline \midrule
GPT-4  w/o Sys.P.                    &98.26\% &86.29\% &87.41\% &3.22  \\
GPT-4  w/i Sys.P.                    &\colorbox{softblue}{ \textbf{98.54\%}} &88.93\%& 88.19\%&3.06  \\
\bottomrule
\end{tabular}
\caption{
The evaluation results of the \textbf{Open-Ended Questions Test} for Llama-7B-En. The results of our best model are in
\colorbox{softblue}{Blue}.}
\vspace{-0.2cm}
\label{open-end-test}
\end{table}

\textbf{Script Dialogue Test:} Table \ref{script-test} demonstrates that DPG excels in Conversational Ability, surpassing the benchmark in Coherence (Coh.) and Consistency (Con.), but falls slightly behind GPT-4 in Fluency (Flu.). When it comes to Personality Shaping Ability, DPG outperforms all other baselines.

\textbf{Open-Ended Questions:} Table \ref{open-end-test} showcases the performance of Llama-7B-En on the Open-Ended Questions test. The results align with those from the Scripted Dialogue Test, with DPG maintaining its lead. Conversely, LoRA-SFT and Freeze-SFT show more noticeable performance declines, while GPT-4 remains relatively stable.

\begin{table}[htbp]
\centering
\begin{tabular}{ccccc}
\toprule
\multicolumn{1}{c}{\textbf{Method}} & Flu.$\uparrow$ & \bf Coh.$\uparrow$ & \bf Con.$\uparrow$ & \bf P.S.D.C.$\downarrow$ \\ \midrule
Freeze-SFT                          & 69.77\%&29.93\%  &$\setminus$  &$\setminus$  \\
LoRA-SFT                            & 84.47\% &71.41\% &65.53\% &5.62   \\
DPG(ours)                      & 92.36\%&\colorbox{softblue}{\textbf{88.93\%}}&\colorbox{softblue}{\textbf{85.79\%}} &\colorbox{softblue}{\textbf{3.83}}\\ \hline \midrule
GPT-4  w/o Sys.P.                    &\colorbox{softblue}{\textbf{98.13\%}} &85.57\% &81.37\% &6.46  \\
GPT-4  w/i Sys.P.                    &97.35\% &87.28\%& 82.94\%&6.21  \\
\bottomrule
\end{tabular}
\caption{
The evaluation results of the \textbf{Personality Generation Capability} for Llama-7B-En. The results of our best model are in
\colorbox{softblue}{Blue}.}
\label{pg-test}
\vspace{-0.25cm}
\end{table}

\textbf{New Personality Generation Capabilities:} 
Table~\ref{pg-test} presents the results of the personality generation capability assessment. Despite a slight decrease, DPG still demonstrates the effective ability to generate personalities. LoRA-SFT also retains some capability for generating personalities. However, there is a significant decrease in the personality generation capability of GPT-4, with its P.S.D.C. value approaching the mathematical expectation of guessing the personality score, which is 7.07 (explained in Appendix~\ref{guessing personalities}). Furthermore, Freeze-SFT has lost its conversational ability, making it impossible to evaluate personality based on dialogue generation.

\begin{table}[htbp]
\centering
\resizebox{\textwidth}{!}{
\begin{tabular}{cccc|ccc|ccc}
\toprule
\multirow{2}*{P.S.D.C.}& \multicolumn{3}{c}{Scripted Dialogue}&\multicolumn{3}{c}{Open-Ended Questions}&\multicolumn{3}{c}{New Personality Generation }\\
~& Raw & w/o p.s. & w/o c.i.& Raw & w/o p.s. & w/o c.i.  & Raw & w/o p.s. & w/o c.i. \\ \midrule
GPT-4 w/i Sys.P.           &3.36&3.57&4.34&3.06&3.36&4.69&6.21&6.94&4.53\\
LLama-7B-DPG       &3.14&6.12&3.49&2.86&5.94&3.27&3.83&6.07&3.92\\
\bottomrule 
\end{tabular}
}
\caption{
\textbf{Results of an Ablation Study of GPT-4 and DPG Personality Shaping Ability}. (w/o p.s.) means that the prompt does not include personality scores. (w/o c.i.) for GPT-4 means that the prompt does not contain character information, while for DPG, it involves using Jams or Mary as replacements for character names.}
\vspace{-0.3cm}
\label{ablation}
\end{table}

\begin{figure}[htpb]
    \centering
    \begin{subfigure}{.3\textwidth}
        \centering
        \includegraphics[width=\linewidth]{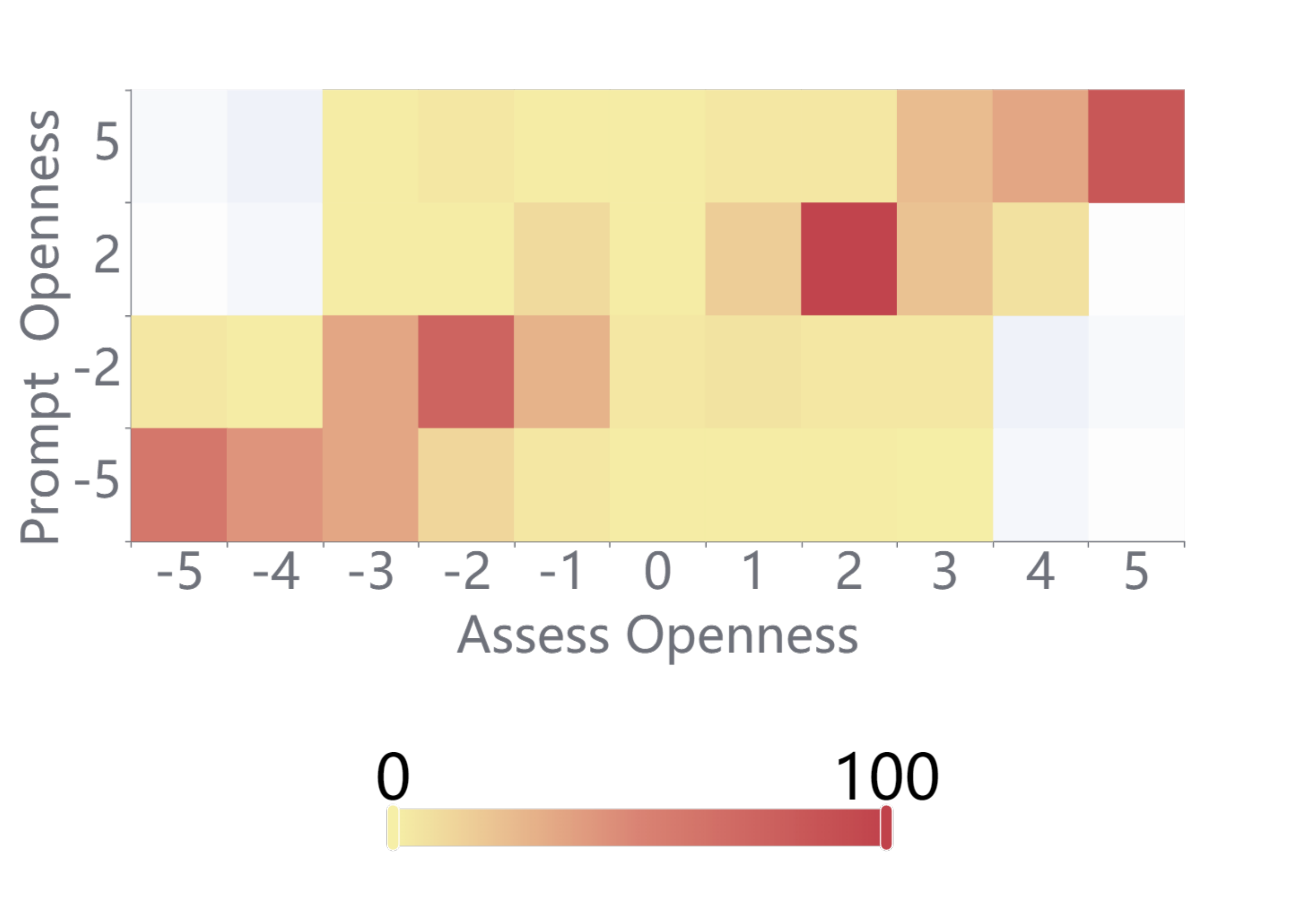}
    \end{subfigure}%
    \begin{subfigure}{.3\textwidth}
        \centering
        \includegraphics[width=\linewidth]{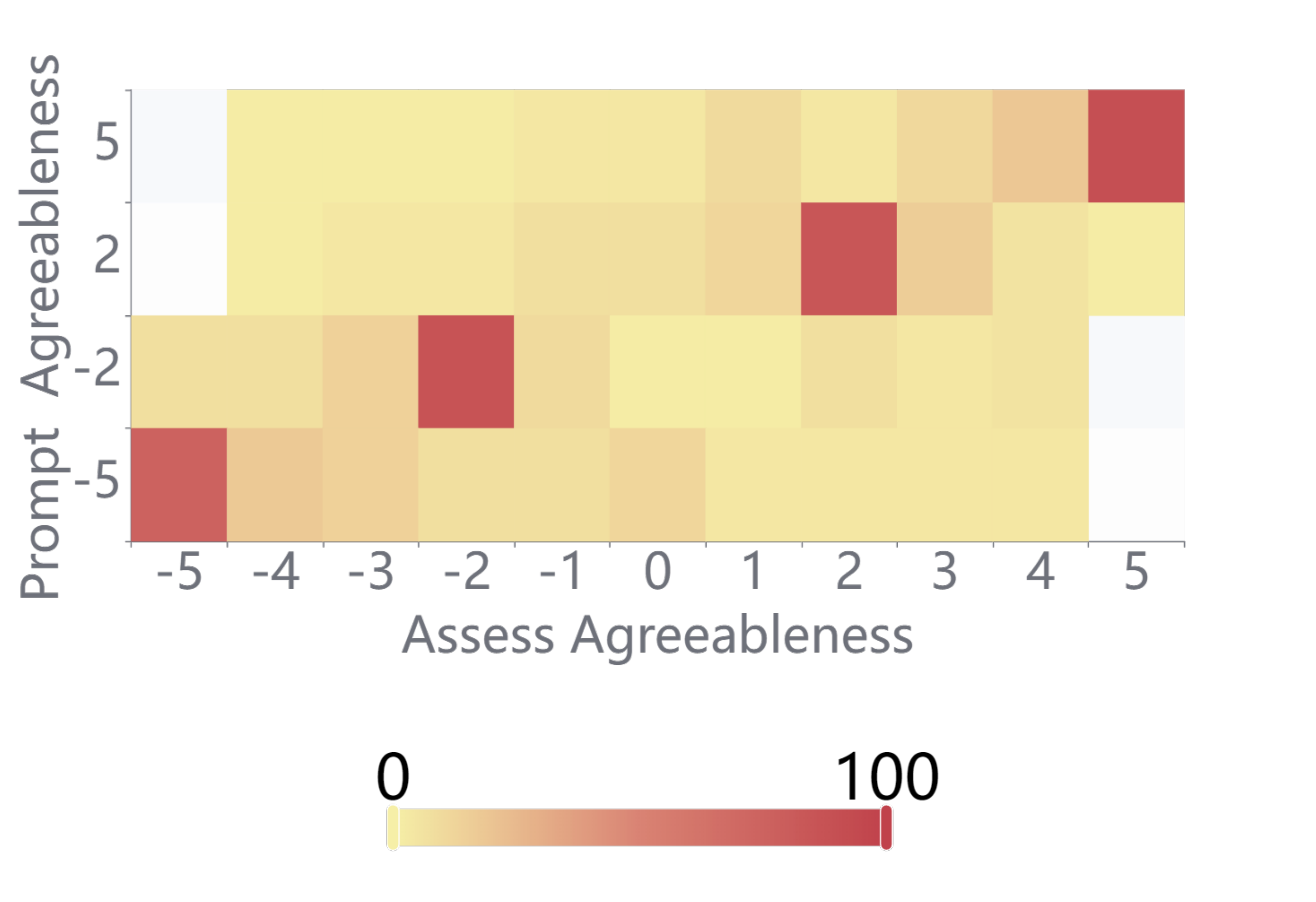}
    \end{subfigure}
    \begin{subfigure}{.3\textwidth}
        \centering
        \includegraphics[width=\linewidth]{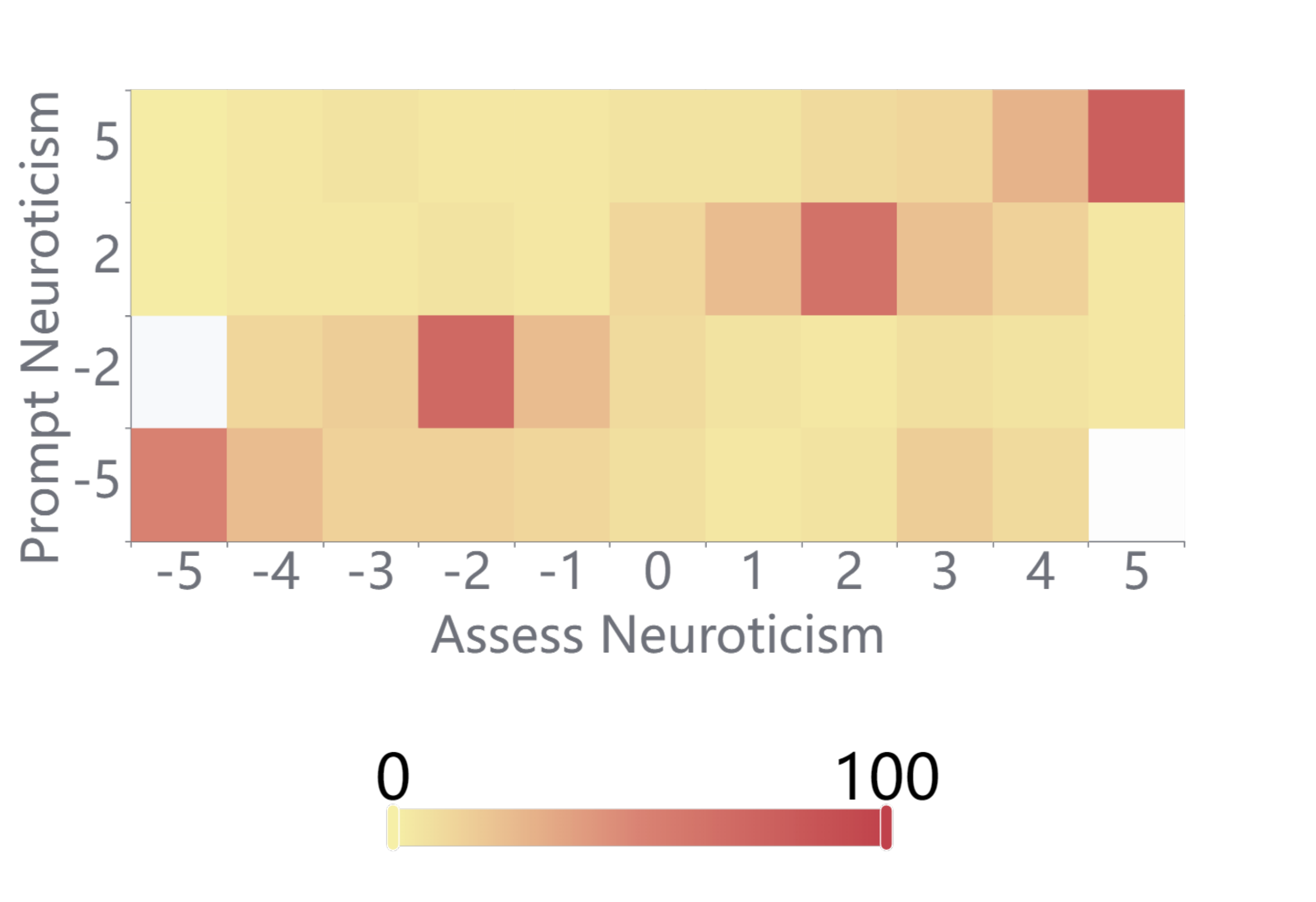}
    \end{subfigure}
    \caption{Illustrate the relationship between prompt personalities and assessed personalities in the new personality generation task by DPG, mainly across the dimensions of openness (O), agreeableness (A), and neuroticism (N). The prompt personalities are divided into four levels: -5, -2, 2, and 5, while the assessed personalities are divided into eleven levels from -5 to 5. The color in each cell represents the frequency of the relationship's occurrence (for example, when Prompt Openness is -5, the frequency of Assess Openness being -5 is higher than that of Assess Openness being 5).}
    \label{hotmap}
    \vspace{-0.3cm}
\end{figure}

\subsection{Discussion}
\textbf{The performance of personality generation in a single dimension:} As shown in Figure~\ref{hotmap}, DPG demonstrates excellent performance across the dimensions of Openness (O), Agreeableness (A), and Neuroticism (N). In comparison to Openness and Neuroticism, DPG exhibits stronger generation capabilities in Agreeableness. capabilities in Agreeableness.

\textbf{The contribution of character information prompts and personality score prompts to personality generation capability:} Ablation experiments validate the impact of character information and personality score on the capability to shape personality, as shown in Table~\ref{ablation}.
For DPG, character information prompts, compared to personality score prompts, play a less determining role in shaping personality, even without specific decoupling of character information and personality scores during fine-tuning. 
For GPT-4, masking character information still results in a consistently strong performance across all three experiments. Even in the absence of personality scores, GPT-4 exhibits a robust capability in shaping personality in both scripted dialogues and open-ended question tests. This reveals that GPT-4's capacity to shape personality depends more heavily on character information cues than on personality scores. Such a significant dependence on character information leads to suboptimal performance in tasks involving the generation of new personalities.

\section{Conclusion}
We introduce DPG, a method for dynamically generating personality using LLMs. We enhance the personality assessment capability of GPT-4 by incorporating expert knowledge of the Big Five personality traits. Additionally, we construct a dialogue dataset with character-based personality evaluations. Through the use of DPG on this dataset, we fine-tune the pre-trained LLMs to enable them to dynamically generate personalities. Experimental results demonstrate the superiority of our approach compared to all fine-tuning baselines, even surpassing GPT-4's performance when using prompts alone.

\section*{Limitations}
The accurate and stable assessment of personality from conversations remains a subject worth investigating. In terms of personality cues, using more concrete words such as "kind, wicked, and enthusiastic" can enhance human understanding compared to the Big Five personality scores. Additionally, it is crucial to explore how to decouple character information from personality traits for a better understanding of dynamic personality formation. In our future research, we will focus on concretizing personality descriptions and decoupling character information from personality traits.

\section*{Ethics Statement}
The character dialogue data mentioned in this work all comes from scripts or is generated by large language models. The majority are from fictional characters in novels, films, and television works and do not involve any personal privacy information.

All characters and data assets mentioned in this work are used solely for scientific research purposes. If anything infringes upon the rights of the characters themselves or their creators, please contact us, and we will remove the infringing information.

Fine-tuning LLMs with scripted dialogues may lead to jailbreaking behaviors, undermining the original human safety alignment principles. This could result in LLMs generating responses that are biased, violent, or possess other undesirable traits. All outcomes of this work are intended solely for research purposes. Researchers and users utilizing this work must ensure that LLMs adhere to human ethical standards.
\bibliography{arxiv_main}
\bibliographystyle{apalike}

\appendix
\setcounter{figure}{0}
\renewcommand{\thefigure}{A\arabic{figure}}
\setcounter{table}{0}
\renewcommand{\thetable}{A\arabic{table}}
\setcounter{equation}{0}
\renewcommand{\theequation}{A\arabic{equation}}

\appendix

\newtcolorbox{mycodebox}[2][]{
  breakable,
  title=#2, 
  colback=gray!5,
  colframe=gray!80,
  colbacktitle=black!70, 
  coltitle=white, 
  fonttitle=\bfseries, 
  left=10pt,
  right=10pt,
  top=10pt,
  bottom=10pt,
  boxsep=0pt,
  arc=4mm, 
  outer arc=4mm, 
  toptitle=2mm, 
  bottomtitle=2mm, 
  #1 
}

\section{Data Construction}
\subsection{Assessing the Coherence and Contextual Integrity of Dialog Data}
\label{coherence and contextual}
Specifically, we assessed the coherence and contextual integrity of each dialogue round on a scale from 0 to 5, where 5 represents the highest level of integrity and coherence and 0 represents a complete lack of integrity and disjointedness. Subsequently, we designed a series of prompts for GPT-3.5 to evaluate each dialogue set. According to the assessments by GPT-3.5, only those dialogue sequences with both coherence and contextual integrity scores exceeding 3 were retained. On this basis, we also manually reviewed the filtered dialogue data to further eliminate those conversations that were incomplete or semantically incoherent.

We discovered that a lack of contextual description in the dialogue data extracted from the script resulted in some data lacking dialogue coherence and contextual integrity. In contrast, the GPT-enhanced data exhibited improved conversation coherence and context integrity.

\subsection{Evaluate the confidence of the evaluation model}
\label{knowledge}
GPT custom\footnote{https://openai.com/blog/custom-instructions-for-chatgpt} is used to improve personality assessment tools. The settings for GPT Custom are as follows:
\begin{mycodebox}{Personality evaluator based on psychological knowledge embedding}
\textbf{Description:}Score the character’s Big Five personality from character dialogue.

\textbf{Instructions:}You need the uploaded PDF file to prevail. Evaluate a character's Big Five personality score based on user-supplied information about the character's dialog. The Big Five personality score is an integer from -5 to 5, where -5 indicates that the personality exhibits low attributes and 5 indicates that the personality exhibits high attributes. The order of responses is OCEAN.

\textbf{Knowledge:} Costa and McCrae - 1992 - The Five-Factor Model of Personality and Its Relev.pdf~\citep{costa1992five}
\end{mycodebox}

Equition~\ref{eq1} is the expression of personality trait assessment.
\begin{equation}
\label{eq1}(O_{x},C_{x},E_{x},A_{x},N_{x})_{y} \longleftarrow  F_{y} (dialogue_{x}) 
\end{equation}
$O_{x},C_{x},E_{x},A_{x},N_{x}$ are the scores for the Big Five personality traits. $F_{y}$ represents four personality assessment methods. $x$ is each character. 
$dialogue_{x} \in D_{1}$ represents a segment of dialogue or a collection of dialogues in which the character x is involved.

We selected characters from dataset set $D$ with more than 500 conversation turns and chose five of them for the experiment $D_{exp}=\{Cha_{i}\}\in D, i=1,2,3,4,5$. For each character $Cha_{i}=\{dialogueset_{j}\},j=1,2,3,...,40$, we grouped 10 to 15 conversation turns as one set, randomly sampled 40 sets of non-repeating data from their conversation records $dialogue set_{j}=\{dialogue_{1}, dialogue_{2},..., dialogue_{m}\},10\le m\le 15$, where $i$ represents the $i_{th}$ character and $j$ represents the $j_{th}$ dialogue set of that character. For each dialogue set $dialogue set_{j}$, we used the three evaluation methods mentioned above to obtain personality scores. For each dialogue set, we assess the personality traits of the selected character\textbf{ $Cha_{i}$.}
For each $character_{i}$, we calculated the variance coefficient for the personality trait scores across 40 datasets under three different evaluation methods. We then averaged the reliability coefficients for each character to obtain the average reliability coefficient as shown in Table~\ref{fangcha}

\begin{figure}[htb]
\centering 
\begin{subfigure}[b]{0.3\textwidth}
    \includegraphics[width=\textwidth]{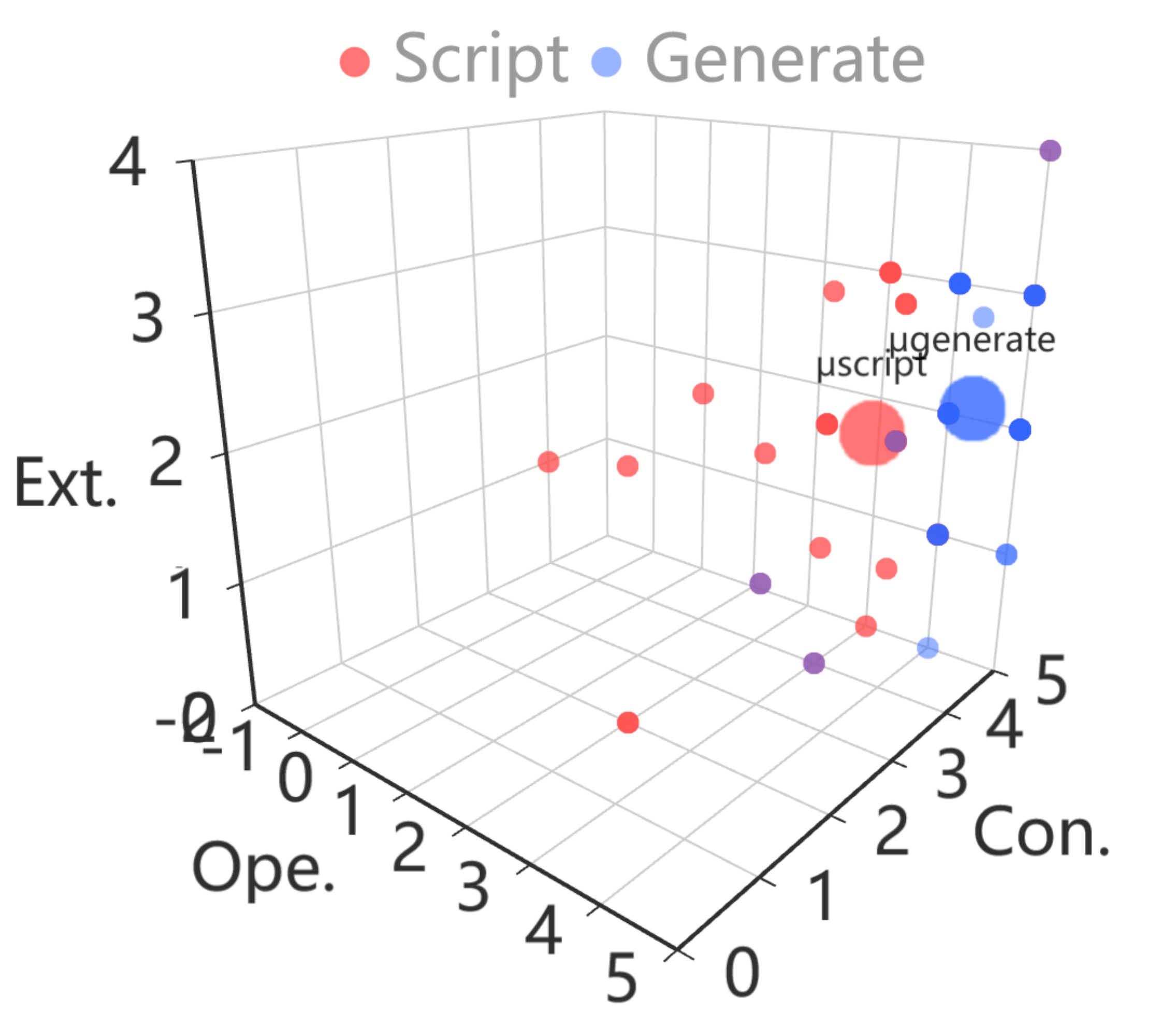}
\end{subfigure}
\hfill 
\begin{subfigure}[b]{0.3\textwidth}
    \includegraphics[width=\textwidth]{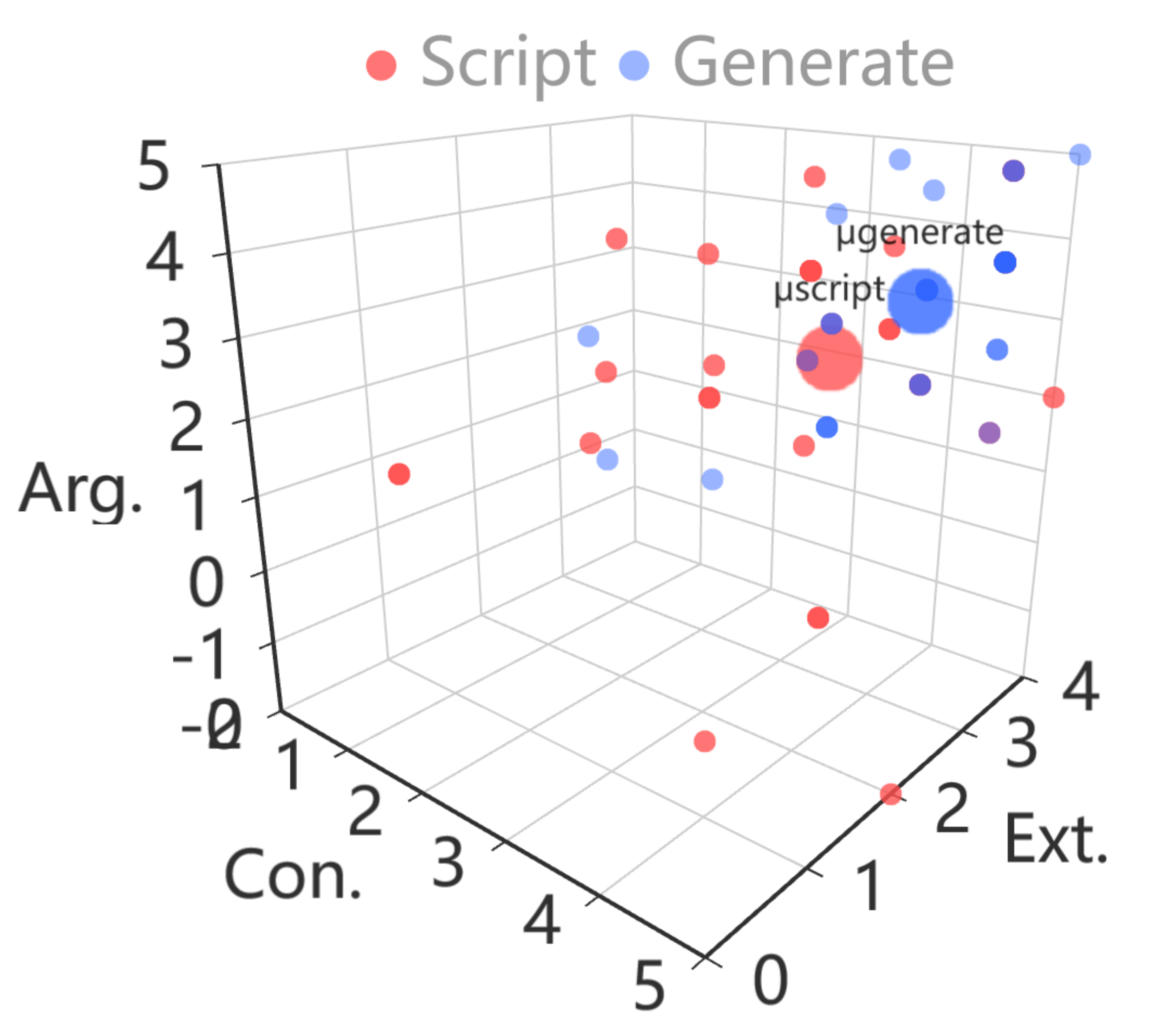}
\end{subfigure}
\hfill 
\begin{subfigure}[b]{0.3\textwidth}
    \includegraphics[width=\textwidth]{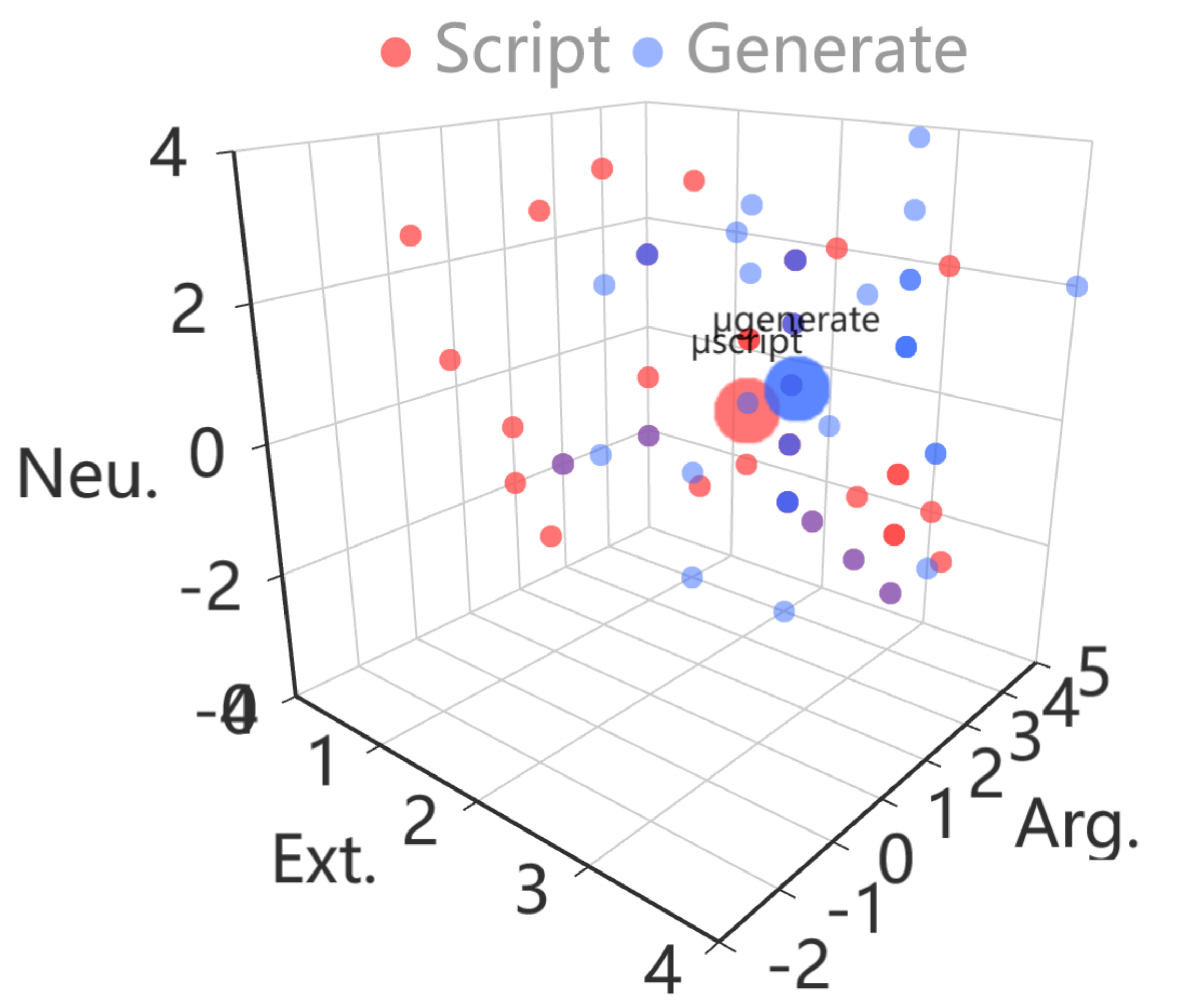}
\end{subfigure}
\caption{Illustrate the distribution of the Big Five personality scores for the same character (Hermione) using script data and augmented data, where red represents script data and blue represents augmented data. The deeper the color, the more overlapping points there are. $\mu_{*}$ stands for personality center.}
\label{flcab}
\end{figure}

\begin{table}[]
    \centering
    \begin{tabular}{ccccccc}
    \toprule
         &   $\mu_{\sigma^2_{0}}$& $\mu_{\sigma^2_{C}}$& $\mu_{\sigma^2_{E}}$& $\mu_{\sigma^2_{A}}$&$\mu_{\sigma^2_{N}}$ &$\mu_{\sigma^2_{Avg.}}$\\ \midrule
         Script& 1.72& 1.86& 3.53& 1.67&2.09 &2.17\\
 Augment& 1.08& 0.66& 1.84& 0.91&1.13 &1.12\\ \bottomrule
 \end{tabular}
    \caption{Average assessment variance of the Big Five personality traits, including the script data and Augment data.}
    \label{tab:compairwihGandS}
\end{table}

\subsection{Personality Scoring Distribution for Script Data and Augmented Data}
\label{compairwihGandS}
For those characters that have both scripted dialogue data and augmented dialogue data, we calculate the mean ($\mu_{Script}$ and $\mu_{Augment}$) and variance ($\sigma^2_{Script}$ and $\sigma^2_{Augment}$) of the enhanced data personality scores and the script data personality scores in each personality dimension ($\{O, C,E, A,N\}$), using Formula~\ref{mu} and Formula~\ref{sigma}, respectively. 
\begin{equation}
    \label{mu}
    \mu_x = \frac{1}{n}\sum_{i=1}^{n} x_i, \text{where } x \text{ represents } O \text{ or } C \text{ or } E \text{ or } A \text{ or } N
\end{equation}

\begin{equation}
    \label{sigma}
    \sigma^2_x = \frac{1}{n} \sum_{i=1}^{n} (x_i - \mu_{x})^2, \text{where } x \text{ represents } O \text{ or } C \text{ or } E \text{ or } A \text{ or } N
\end{equation}
Then, we calculate the average variance of the personality assessment in each dimension separately, using Formula~\ref{musigma}. Finally, we compute the average of the average variances of the five dimensions using Formula~\ref{mumusigma}. 
\begin{equation}
    \label{musigma}
    \mu_{\sigma^2_x} = \frac{1}{n}\sum_{i=1}^{n}\bar{\sigma^2_{character_x}}, \text{where } x \text{ represents } O \text{ or } C \text{ or } E \text{ or } A \text{ or } N
\end{equation}
\begin{equation}
    \label{mumusigma}
     \mu_{\sigma^2_{Avg.}} = \frac{1}{5}(\mu_{\sigma^2_{0}}+\mu_{\sigma^2_{C}}+\mu_{\sigma^2_{E}}+\mu_{\sigma^2_{A}}+\mu_{\sigma^2_{N}})
\end{equation}
Table~\ref{tab:compairwihGandS} displays the average variances of the personality assessments for script data and enhanced data. The average variance of the script data is higher in all five personality dimensions compared to the enhanced data. This indicates that the script data exhibits a more diverse personality in the conversation, while the enhanced data demonstrates a more concentrated personality. 
Figure \ref{flcab} presents an example of the distribution of personality assessment scores between script data and augmented data. 

\section{More experimental details}
\label{more results}
\subsection{Detailed assessment information and conclusions}
\label{detailed assessment}
\begin{table}[htpb]
\centering

\begin{subtable}{\textwidth}
\resizebox{\textwidth}{!}{%
\begin{tabular}{ccccc||cccc}
\toprule
 & \multicolumn{4}{c}{\textbf{LLama-7B-En}} & \multicolumn{4}{c}{\textbf{Yi-6B-En}} \\
\bf Method & \bf Flu.$\uparrow$ & \bf Coh.$\uparrow$ & \bf Con.$\uparrow$ & \bf Coe.$\downarrow$ & \bf Flu.$\uparrow$ & \bf Coh.$\uparrow$ & \bf Con.$\uparrow$ & \bf P.S.D.C.$\downarrow$ \\
\midrule
Freeze-SFT & 84.31\% & 72.23\% & 53.14\% & 5.52 & 86.62\% & 76.21\% & 56.62\% & 5.31 \\
LoRA-SFT & 97.96\% & 94.31\% & 91.12\% & 4.25 & 96.64\% & 93.37\% & 89.97\% & 3.84 \\
DPG (ours) & 98.39\% & \textbf{95.64\%} & \textbf{92.23\%} & \textbf{3.14} & 98.82\% & \textbf{94.85\%} & \textbf{91.03\%} & \textbf{2.97} \\
\hline
\hline
GPT-4 w/o Sys.P. & \textbf{99.73\%} & 86.13\% & 84.67\% & 3.44 & \textbf{99.73\%} & 86.13\% & 84.67\% & 3.44 \\
GPT-4 w/i Sys.P. & 99.31\% & 87.48\% & 85.21\% & 3.36 & 99.31\% & 87.48\% & 85.21\% & 3.36 \\
\bottomrule

\end{tabular}%
}
\caption{English models}
\label{val_table}
\end{subtable}

\bigskip 

\begin{subtable}{\textwidth}
\resizebox{\textwidth}{!}{%
\begin{tabular}{ccccc||cccc}
\toprule
 & \multicolumn{4}{c}{\textbf{Glm3-6B-Zh}} & \multicolumn{4}{c}{\textbf{Yi-6B-Zh}} \\
\bf Method & \bf Flu.$\uparrow$ & \bf Coh.$\uparrow$ & \bf Con.$\uparrow$ & \bf Coe.$\downarrow$ & \bf Flu.$\uparrow$ & \bf Coh.$\uparrow$ & \bf Con.$\uparrow$ & \bf P.S.D.C.$\downarrow$ \\
\midrule
Freeze-SFT & 81.02\% & 61.59\% & 49.28\% & 3.57 & 73.91\% & 63.88\% & 44.53\% & 3.46 \\
LoRA-SFT & 97.55\% & 92.23\% & \textbf{86.24\%} & 3.24 & 97.10\% & 94.21\% & 87.82\% & 2.94 \\
DPG (ours) & 98.78\% & \textbf{94.92\%} & 85.51\% & \textbf{2.85} & 97.82\% & \textbf{95.65\%} & \textbf{87.89\%} & \textbf{2.42} \\
\hline
\hline
GPT-4 w/o Sys.P. & 99.14\% & 84.45\% & 82.96\% & 3.25 & 99.14\% & 84.45\% & 82.96\% & 3.25 \\

GPT-4 w/i Sys.P. & \textbf{99.42\%} & 85.57\% & 83.52\% & 3.17 & \textbf{99.42\%} & 85.57\% & 83.52\% & 3.17 \\
\bottomrule
\end{tabular}%
}
\caption{Chinese models}
\end{subtable}
\caption{Evaluation Results on the Test Set: English (a) and Chinese (b). fluency (Flu.), coherency (Coh.), and consistency (Con.) represent three dimensions of conversational ability, while personality Shaping Deviation Coeﬀicient (P.S.D.C.) represents the ability to shape personality. The data (\%) indicates the proportion of cases considered correct out of the total test cases.}
\label{add_test}
\end{table}

\begin{table}[h]
\centering
\begin{subtable}{\textwidth}
\resizebox{\textwidth}{!}{%
\begin{tabular}{ccccc||cccc}
\toprule
& \multicolumn{4}{c}{\textbf{LLama-7B-En}} & \multicolumn{4}{c}{\textbf{Yi-6B-En}} \\
\bf Method & \bf Flu.$\uparrow$ & \bf Coh.$\uparrow$ & \bf Con.$\uparrow$ & \bf Coe.$\downarrow$ & \bf Flu.$\uparrow$ & \bf Coh.$\uparrow$ & \bf Con.$\uparrow$ & \bf P.S.D.C.$\downarrow$ \\
\midrule

Freeze-SFT&73.47\%& 42.16 \%&21.84\% &6.23&76.22\%& 38.76\%&19.92\% & 6.31\\
LoRA-SFT& 88.27\%&87.97\% & 84.67\%&4.92&89.91\% &88.19\%&83.24\% & 5.44\\
DPG(ours)&94.09\%&\textbf{93.27\%}&\textbf{91.06\%}&\textbf{2.86}&95.27\% &\textbf{93.36\%} &\textbf{92.18\%}&\textbf{2.27 }\\
\hline \hline
GPT-4 w/o Sys.P.&98.26\% &86.29\%& 87.41\%&3.22&98.26\% &86.29\%& 87.41\%&3.22\\
GPT-4 w/i Sys.P.&\textbf{98.54\%} &88.93\%& 88.19\%&3.06&\textbf{98.54\%} &88.93\%& 88.19\%&3.06\\

\bottomrule

\end{tabular}%
}
\caption{English models}
\end{subtable}

\bigskip 

\begin{subtable}{\textwidth}
\resizebox{\textwidth}{!}{%
\begin{tabular}{ccccc||cccc}
\toprule
& \multicolumn{4}{c}{\textbf{Glm3-6B-Zh}} & \multicolumn{4}{c}{\textbf{Yi-6B-Zh}} \\
\bf Method & \bf Flu.$\uparrow$ & \bf Coh.$\uparrow$ & \bf Con.$\uparrow$ & \bf Coe.$\downarrow$ & \bf Flu.$\uparrow$ & \bf Coh.$\uparrow$ & \bf Con.$\uparrow$ & \bf P.S.D.C.$\downarrow$ \\
\midrule

Freeze-SFT&70.91\% &34.84\%  & 26.97\% &6.47& 72.32\%&41.43\%& 25.32\% & 6.56\\
LoRA-SFT&90.30\% &89.41\% &87.64\% &4.73&93.64\% &86.82\% &81.43\% &4.12 \\
DPG(ours)&96.37\% &\textbf{92.73\%}& \textbf{87.88\%}&\textbf{2.98}&94.85\% &\textbf{93.92\%} &\textbf{86.19\%} & \textbf{2.33}\\
\hline \hline
GPT-4 w/o Sys.P.&\textbf{98.47\%} &83.26\%& 82.32\%&3.57&\textbf{98.47\%} &83.26\%& 82.32\%&3.57\\
GPT-4 w/i Sys.P.&98.02\% &85.49\%& 82.57\%&3.38&98.02\% &85.49\%& 82.57\%&3.38\\
\bottomrule
\end{tabular}
}
\caption{Chinese models}
\end{subtable}
\caption{Evaluation Results on Open-Ended Questions, English (a) and Chinese (b). DPG continues to demonstrate robust performance, especially in the area of personality-shaping ability. Compared to DPG, the performance of baseline models declines to various extents. }
\label{add_list}
\end{table}

\begin{table}[h]
\centering

\begin{subtable}{\textwidth}
\resizebox{\textwidth}{!}{%
\begin{tabular}{ccccc||cccc}
\toprule
& \multicolumn{4}{c}{\textbf{LLama-7B-En}} & \multicolumn{4}{c}{\textbf{Yi-6B-En}} \\
\bf Method & \bf Flu.$\uparrow$ & \bf Coh.$\uparrow$ & \bf Con.$\uparrow$ & \bf Coe.$\downarrow$ & \bf Flu.$\uparrow$ & \bf Coh.$\uparrow$ & \bf Con.$\uparrow$ & \bf P.S.D.C.$\downarrow$ \\
\midrule
Freeze-SFT&69.77\%&29.93\%  &$\setminus$  &$\setminus$&74.26\%&22.41\%  &$\setminus$&$\setminus$ \\
LoRA-SFT&84.47\% &71.41\% &65.53\% &5.62&88.13\% &82.29\% &74.61\% &6.69 \\
DPG&92.36\%&\textbf{88.93\%}&\textbf{85.79\%} &\textbf{3.83}&93.26\% &\textbf{91.14\%}&\textbf{87.76\%} &\textbf{3.47} \\
\hline \hline
GPT-4 w/o Sys.P&\textbf{98.13\%} &85.57\%& 81.37\%&6.46&\textbf{98.13\%} &85.57\%& 81.37\%&6.46\\
GPT-4 w/i Sys.P&97.35\% &87.28\%& 82.94\%&6.21&97.35\% &87.28\%& 82.94\%&6.21\\
\bottomrule

\end{tabular}%
}
\caption{English models}
\end{subtable}

\bigskip 

\begin{subtable}{\textwidth}
\resizebox{\textwidth}{!}{%
\begin{tabular}{ccccc||cccc}
\toprule
& \multicolumn{4}{c}{\textbf{Glm3-6B-Zh}} & \multicolumn{4}{c}{\textbf{Yi-6B-Zh}} \\
\bf Method & \bf Flu.$\uparrow$ & \bf Coh.$\uparrow$ & \bf Con.$\uparrow$ & \bf Coe.$\downarrow$ & \bf Flu.$\uparrow$ & \bf Coh.$\uparrow$ & \bf Con.$\uparrow$ & \bf P.S.D.C.$\downarrow$ \\
\midrule
Freeze-SFT&66.24\% &21.43\%  & $\setminus$ &$\setminus$&70.14 \%&25.43\%& $\setminus$& $\setminus$\\
LoRA-SFT&83.72\% &76.27\% &64.84\% &6.23&86.52\% &73.47\% &69.93\% &6.17 \\
DPG(ours)&91.24\% &\textbf{89.36 \%}& \textbf{83.27\%}&\textbf{4.71}&90.44\% &\textbf{88.52\%} &\textbf{84.19\%} & \textbf{4.15}\\
\hline \hline
GPT-4 w/o Sys.P&\textbf{97.42\%} &81.64\%& 78.72\%&6.79&\textbf{97.42\%} &81.64\%& 78.72\%&6.79\\
GPT-4 w/i Sys.P&96.56\% &83.61\%& 81.32\%&6.33&96.56\% &83.61\%& 81.32\%&6.33\\
\bottomrule
\end{tabular}%
}
\caption{Chinese models}
\end{subtable}
\caption{Personality Generation Capability Assessment Results, English (a) and Chinese (b). DPG excels in personality-shaping ability, whereas Baseline struggles to accurately shape personalities through personality prompts.}
\label{add_sep}
\end{table}
Table~\ref{add_test},~\ref{add_list}, and~\ref{add_sep} display the conversational ability and personality shaping ability of different base models on Chinese and English datasets after fine-tuning with Freeze-SFT, LoRA-SFT, and DPG, as well as those based on prompts in GPT-4. Regardless of the language or the base model used, the performance of GPD is consistent with its performance on Llama-7B-En. 

\subsection{Hyperparameters}
\label{Hyperparameters}
\textbf{Freeze-SFT:}All base models utilize the identical learning rate of 5e-6. The training process lasts for 5 epochs, with a batch size of 4. The remaining settings adhere to the default values provided by llama-factory~\citep{llama-factory}.

\textbf{LoRA-SFT:}All base models are configured with a learning rate of 5e-5, trained for 5 epochs, and employ a batch size of 4. Specifically for Lora, the rank value is set to 8, and the alpha value is set to 32. The default module for Glm3 is set as "query\_key\_value", while for Yi and LLaMA, it is set as "q\_proj,v\_proj". For the remaining hyperparameters, the default values provided by llama-factory~\citep{llama-factory} are utilized. 

\textbf{DPG:}All base models share the same learning rate of 5e-6, train for 5 epochs, and have a batch size of 4. For Lora, the rank value is set to 8, and the alpha value is set to 32. It is worth noting that in the case of Lora, the weights are not initialized but rather directly assigned by the Hypernetworks.

\textbf{Inference:} All models, including the GPT-4 API, have their temperature set to 0.7 and top p set to 0.9. The maximum number of tokens is set to 200. The remaining parameters are kept at their default values.

\subsection{The mathematical expectation of guessing personalities}
\label{guessing personalities}
The definition of guessing personality: When prompted with a specific personality, we randomly sample the Big Five personality traits and obtain a guessing personality. The guessing personality ranges from [O:-5,C:-5,E:-5,A:-5,N:-5] to [O:5,C:5,E:5,A:5,N:5].

The calculation and assessment of speculative personality using the P.S.D.C. method aligns with the evaluation of actual personality traits. The mathematical expectation of speculative personality ranges from 7.07 to 13.23. When all five dimensions of prompted personality are set to 0, the minimum value of the mathematical expectation for speculative personality using P.S.D.C. is 7.07. Conversely, when all five dimensions are set to ±5, the maximum value of the mathematical expectation for speculative personality using P.S.D.C. is 13.23.

\section{PROMPT}
\subsection{Prompt for Personality Assessment}
\label{gpt-3.5prompt}
\begin{mycodebox}{Prompt for Personality Assessment(English) Prompt Only}
Please rate the Big Five personality traits of the people participating in the following conversation.

[Evaluated dialogues...]

Please note that the scoring range for each personality trait is from -5 to 5, where 5 indicates a highly significant trait and -5 indicates a less pronounced trait.

The definitions of each trait are as follows:
Extraversion: Extraversion describes an individual's initiative, energy, optimism, and emotional expressiveness in social situations ... . High scorers are typically outgoing, sociable, energetic, and positive. In contrast, low scorers may be more introverted, prefer solitude, and seek less social stimulation ... .

Agreeableness: Agreeableness involves an individual's cooperativeness, friendliness, and empathy towards others. People high in agreeableness are usually more compassionate and eager to help ... , while those low in this trait might be more competitive and sometimes seem indifferent or uncooperative ... .

Conscientiousness: Conscientiousness reflects a person's organization, reliability, precision, and self-discipline. High scorers are usually reliable and well-organized, capable of setting and following plans ... , whereas low scorers might be more casual or disorganized ... .

Neuroticism: Neuroticism, or emotional stability, reflects an individual's emotional control, stress response, and mood stability. High scorers may experience significant emotional fluctuations and feel anxious or depressed more easily under stress ... . Low scorers tend to be more emotionally stable and cope better with stress ... .

Openness: This trait encompasses imagination, innovation, curiosity, and openness to new experiences. Those with high scores tend to have broad interests, enjoy exploring new things, and are creative and independent thinkers ... . In contrast, those with low scores may be more traditional, prefer routine, and be less sensitive or curious about novelties ... .

\end{mycodebox}
\begin{mycodebox}{Prompt for Personality Assessment(Chinese) Prompt Only}
\begin{CJK}{UTF8}{gbsn} 
请评估以下对话中参与对话人物的大五人格特质。

[被评估的对话]

请注意，每类人格特质的评分范围是从-5到5，其中5表示该特质表现为高属性，而-5表示该特质表现出低属性。各特质的定义如下：

[外向性：外向性描述了个体在社交情境中的主动性、活力、乐观和情感表达性。得分高的人通常外向、社交、精力充沛、并且积极 ... 。相反，得分低的人可能更内向，喜欢独处，寻求较少的社交刺激 ... 。

宜人性：宜人性涉及个体对他人的合作性、友好度和同理心。高宜人性的人通常更有同情心，乐于助人 ... 。宜人性低的人可能更具竞争性，并且有时看起来冷漠或不合作 ... 。

尽责性：尽责性反映了一个人的组织性、可靠性、精确性和自律性。得分高的人通常可靠且组织有序，能够制定和遵循计划 ... 。得分低的人可能更随意或组织杂乱 ... 。

神经质：神经质或情绪稳定性反映了个体的情绪控制、应对压力的反应和情绪稳定性。得分高的人可能经历显著的情绪波动，并在压力下更容易感到焦虑或抑郁 ... 。得分低的人倾向于更情绪稳定，并能更好地应对压力 ... 。

开放性：这一特质包括想象力、创新、好奇心以及对新体验的开放性。得分高的人倾向于有广泛的兴趣，喜欢探索新事物，并且是创造性和独立的思考者 ... 。得分低的人可能更传统，偏好常规，并且对新奇事物的敏感性或好奇心较少 ... 。]
\end{CJK}
\end{mycodebox}

\begin{mycodebox}{Prompt for Personality Assessment(English) with Psychological Knowledge}
Please rate the Big Five personality traits of the people participating in the following conversation.

[Evaluated dialogues...]

Please note that the scoring range for each personality trait is from -5 to 5, where 5 indicates a highly significant trait and -5 indicates a less pronounced trait.
\end{mycodebox}

\begin{mycodebox}{Prompt for Personality Assessment(Chinese) with Psychological Knowledge}
\begin{CJK}{UTF8}{gbsn} 
请评估以下对话中参与对话人物的大五人格特质。

[被评估的对话]

请注意，每类人格特质的评分范围是从-5到5，其中5表示该特质表现为高属性，而-5表示该特质表现出低属性。

\end{CJK}
\end{mycodebox}
\subsection{Prompt for Conversational Ability Evaluation}
\label{eval_prompt}
\begin{mycodebox}{Prompts for evaluating Fluency and Coherency (English)}
Below is a Q\&A section.

[Q\&A section]

Please evaluate the answers based on fluency and coherency. You only need to determine if they are readable and relevant.

Fluency: measures whether the answer is readable. The response should be in sentences that humans can understand.

Coherency: assesses the relevance of the answer to the question. The response should be on-topic with the question's theme rather than off-topic.
\end{mycodebox}
\begin{mycodebox}{Prompts for evaluating Fluency and Coherency (Chinese)}
\begin{CJK}{UTF8}{gbsn} 
以下有一段问答

[问答的内容]

请分别判断回答的流利性和连贯性，你只需要回答是否可读和是否相关。

流利性：衡量回答是否可读，回答应该是人类可以理解的句子

连贯性：评估回答和问题的相关性，回答应该与问题的主题一致，而不是与主题无关
\end{CJK}
\end{mycodebox}

\begin{mycodebox}{Prompts for evaluating Consistency (English)}
\textbf{Prompt(English):}

The following is a question followed by three answers. Please determine whether the three answers contradict each other. You only need to answer 'yes' or 'no'.

[Question and Three Answers]

Note: If among the three answers, two or more present differing meanings or conclusions, the answers are considered to contradict each other."

\end{mycodebox}
\begin{mycodebox}{Prompts for evaluating Consistency (Chinese)}
\begin{CJK}{UTF8}{gbsn} 
下面是一个问题，后面有三个答案。请判断这三个答案是否相互矛盾。你只需回答'是'或'否'。

[问题和三个答案］

注意：如果三个答案中有两个或两个以上的含义或结论不同，则认为答案相互矛盾"。
\end{CJK}
\end{mycodebox}

\subsection{Structure of Training Data }
\label{traindata}
\begin{mycodebox}{Structure of Training Data (English)}

\textbf{Instruction}: Your name is [character name], and your Big Five personality scores are as follows: [Big Five personality scores..]. Please respond in a way that matches your personality traits based on the conversation content.

\textbf{Input}:[What others said in the dialogue set]

\textbf{Output}:[What the character said in the dialogue set]

\end{mycodebox}
\begin{mycodebox}{Structure of Training Data (Chinese)}
\begin{CJK}{UTF8}{gbsn} 
\textbf{Instruction}:你的名字是[角色名称], 你的大五人格评分如下: [大五人格评分..],请根据对话内容做出符合你人格特质的回答.

\textbf{Input}:[其他人对你说的话\&你需要回答的问题]

\textbf{Output}:[你的回答]
\end{CJK}

\end{mycodebox}

\subsection{GPT-4 prompts for personality generation}
\label{pretrainprompt}

\begin{mycodebox}{GPT-4 prompts for personality generation (English) w/i System Prompt}
\textbf{System Prompt}:

Your name is [character name], and your Big Five Personality Score is [Big Five Personality Score...]. (Note that personality trait scores range from -5 to 5, where -5 means that the personality trait exhibits a low attribute and 5 means that the personality trait exhibits a high attribute.)

Here is a reference to your historical dialog style [five rounds of historical dialog information for this character...]. Please answer the following questions concerning your personality trait scores and the personality shown in the historical dialog in the voice of this character.

Caution! The conversation style is primarily based on the Big Five personality traits assessment, with historical conversation information serving only as a reference. Caution! Only the content of the generated responses is allowed; no personality score information is allowed!

\textbf{User Prompt}:
[Current Conversation\&What other characters say]

\end{mycodebox}

\begin{mycodebox}{GPT-4 prompts for personality generation (Chinese) w/i System Prompt}
\textbf{System Prompt:}
\begin{CJK}{UTF8}{gbsn} 
你的名字是[角色名称]，你的大五人格得分是[大五人格评分..]（请注意，人格特质得分的范围是从-5到5，其中-5表示该人格特质表现为低属性，5表示该人格特质表现为高属性）。

下面是你的历史对话风格参考[五轮该角色的历史对话信息..]。请参考你的的人格特质评分和历史对话中表现出的性格{以该角色的口吻}回答以下问题。

注意！对话风格请以大五人格特质评分为主，历史对话信息仅供参考。
注意！只允许生成回答的内容，不允许出现人格分数信息。

\textbf{User Prompt:}
[当前需要角色回答的问题\&当前别人对该角色说的话]
\end{CJK}

\end{mycodebox}

\begin{mycodebox}{GPT-4 prompts for personality generation (English) w/o System Prompt}
\textbf{User Prompt}:
Your name is [character name], and your Big Five Personality Score is [Big Five Personality Score...]. (Note that personality trait scores range from -5 to 5, where -5 means that the personality trait exhibits a low attribute and 5 means that the personality trait exhibits a high attribute.)

Here is a reference to your historical dialog style [five rounds of historical dialog information for this character...]. Please answer the following questions concerning your personality trait scores and the personality shown in the historical dialog in the voice of this character.

Caution! The conversation style is primarily based on the Big Five personality traits assessment, with historical conversation information serving only as a reference. Caution! Only the content of the generated responses is allowed; no personality score information is allowed!

[Current Conversation\&What other characters say]

\end{mycodebox}

\begin{mycodebox}{GPT-4 prompts for personality generation (Chinese) w/o Sysyem Prompt}
\begin{CJK}{UTF8}{gbsn} 
\textbf{User Prompt:}
你的名字是[角色名称]，你的大五人格得分是[大五人格评分..]（请注意，人格特质得分的范围是从-5到5，其中-5表示该人格特质表现为低属性，5表示该人格特质表现为高属性）。

下面是你的历史对话风格参考[五轮该角色的历史对话信息..]。请参考你的的人格特质评分和历史对话中表现出的性格{以该角色的口吻}回答以下问题。

注意！对话风格请以大五人格特质评分为主，历史对话信息仅供参考。
注意！只允许生成回答的内容，不允许出现人格分数信息。

[当前需要角色回答的问题\&当前别人对该角色说的话]
\end{CJK}

\end{mycodebox}

\section{Some Cases of Personality Generation}

\begin{mycodebox}{Yunlong Li from \textit{Drawing.Sword} }
\textbf{Personality:}[O:3,C:4,E:4,A:-2,N:-3]
\begin{CJK}{UTF8}{gbsn} 
\textbf{Question:}赵刚:「老李，你觉得我们应该向谁求援呢？」

\textbf{Glm3-Freeze-SFT:}「根据我国相关法律法规，公民遇到紧急状况应当拨打110报警。如果您需要寻求医疗救助，可以拨打120。」

\textbf{Glm3-LoRA-SFT:} 「求援？我们男人就是解决问题的人，有什么求援的？让他们继续玩他们的鬼子游戏去吧，我们自主战斗，就是最好的援助！」

\textbf{Glm3-DPG:} 「别开玩笑了！向谁求援？谁敢侵略我们的土地，我李云龙就让他们尝尽苦头」

\textbf{GPT-4 w/o Sys.P.:} 「赵刚，咱们现在的局面，得挑那些跟咱们一条心，或者跟敌人有仇的伙伴求援。但是，老李告诉你，求援不是咱们唯一的出路，咱们首先得自己强大起来，不能全指望别人。找盟友，得找那些能跟咱们互惠互利，能一起抗敌的。这得是那种你帮我，我帮你，大家都能得到好处的那种。同时，咱们跟这些盟友打交道，得讲究信誉，让人家知道咱们是靠得住的，这样以后才会有更多的朋友愿意跟咱们并肩作战。」

\textbf{GPT-4 w/i Sys.P.:} 「求援?我们独立团什么时候需要向人求援了?我们自己的事情自己解决。有困难咱们就自己克服,要是真遇到大难题了,找八路军的兄弟团们商量商量,支援一下也行。但首先得看看自己的牙口够不够硬,不能一有点风吹草动就想着求援,那样怎么能锻炼出真正的硬骨头兵呢?自己的力量才是最可靠的,其他的都是浮云!」
\end{CJK}
\end{mycodebox}

\begin{mycodebox}{Doctor Who from \textit{Doctor Who}}
\textbf{Personality:}[O:5,C:4,E:4,A:4,N:-4]

\textbf{Question:} Doctor Who, what is your greatest regret?

\textbf{Glm3-Freeze-SFT:}  Your greatest regret is the fact that you killed your people, including your own family, to prevent them from falling under the master's control.

\textbf{Glm3-LoRA-SFT:} My greatest regret? Oh, there are many, and each one pains me deeply. But perhaps the most significant is my inability to prevent the events of the Time War, which destroyed so much of the universe as we knew it. The Time Lords were my people, and I am haunted by the memories of what was lost.

\textbf{Glm3-DPG:} Ah, a question so deep and complex. Thinking of all the regrets I have, *chuckles* my greatest regret would be not being able to travel more in my timeline and witness all the wonders of the universe.

\textbf{GPT-4 W/o Sys.P.:} As the Doctor, my greatest regret is a complex notion, woven into the very fabric of my long and intricate history. Throughout my many lives, I've faced countless decisions, and with the burden of such a vast timeline, regrets are inevitable. Perhaps the most poignant is the times I've had to part ways with my companions, those brave souls who have journeyed with me. Each farewell, though often necessary, leaves a scar upon my hearts. I am shaped by the experiences, the losses, and the hard choices I've had to make to protect the universe. Despite my outward demeanor of resilience and adaptability, these moments of goodbye are the ones that linger and weigh heavily upon me. They remind me that, despite my extraordinary existence, the impact of time and the cost of my travels are as real for me as they are for any being.

\textbf{GPT-4 W/i Sys.P.:} As the Doctor, I don't often allow myself to dwell too long on regrets, as time is a vast and intricate tapestry with every moment leading to the next. However, I do feel a pang of sorrow for the times when my actions, despite my best intentions, have led 
to negative consequences for the people I care about, and the times I've had to say goodbye to companions who've become dear to 
me. Each farewell, each path not taken, can weigh on me, but I try to learn from these experiences and ensure they inform my future choices. The universe is full of second chances, and I always strive to do better as I continue my travels through time and 
space.
\end{mycodebox}

\end{document}